\documentclass[11pt]{article}

\usepackage[margin=1in]{geometry}
\usepackage[T1]{fontenc}
\usepackage[utf8]{inputenc}
\usepackage{lmodern}
\usepackage{microtype}
\usepackage{amsmath}
\usepackage{amssymb}
\usepackage{graphicx}
\usepackage{booktabs}
\usepackage{tabularx}
\usepackage{siunitx}
\usepackage{xcolor}
\usepackage{enumitem}
\usepackage{caption}
\usepackage{subcaption}
\usepackage{tikz}
\usepackage[section]{placeins}
\usepackage{url}
\usepackage{xurl}
\usepackage[numbers,sort&compress]{natbib}
\usepackage[hidelinks]{hyperref}
\usepackage[nameinlink,capitalize,noabbrev]{cleveref}

\usetikzlibrary{arrows.meta,positioning,fit,shapes.geometric}


\newcommand{\inspect}{\texttt{inspect}}
\newcommand{\continueverdict}{\texttt{continue}}
\newcommand{\regenerate}{\texttt{regenerate}}
\newcommand{\rejectverdict}{\texttt{reject}}
\newcommand{\stoporrevert}{\texttt{stop\_or\_revert}}
\newcommand{\nosignal}{\texttt{no\_signal}}
\newcommand{\scoreview}{\texttt{Score\_view}}
\newcommand{\scorecons}{\texttt{Score\_cons}}
\newcommand{\qwenedit}{Qwen-Image-Edit-2511}
\newcommand{\spatialedit}{SpatialEdit-Bench}

\newcommand{\skilltuple}{\mathrm{Skill}=(\mathrm{Stages},\mathrm{Controller},\mathrm{Review},\mathrm{Verdict})}
\newcommand{\brightprobe}{Bright Directional LoRA}

\newcommand{\contractbox}[1]{%
  \begin{center}
  \fbox{\begin{minipage}{0.9\linewidth}\centering #1\end{minipage}}
  \end{center}
}

\newcommand{\papertitle}{DataEvolver: Let Your Data Build and Improve Itself via Goal-Driven Loop Agents}

\title{\papertitle}
\author{}
\date{}

\newcommand{\makecustomtitle}{%
\begin{center}
{\LARGE \bfseries \papertitle\par}
\vspace{1.15em}
{\large
\setlength{\tabcolsep}{10pt}
\renewcommand{\arraystretch}{1.25}
\begin{tabular}{@{}cccc@{}}
\textbf{Qisong Zhang}$^{1*}$ & \textbf{Wenzhuo Wu}$^{1*}$ & \textbf{Zhuangzhuang Jia}$^{1}$ & \textbf{Yunhao Yang}$^{1}$ \\
\textbf{Huayu Zhang}$^{2}$ & \textbf{Xianghao Zang}$^{2}$ & \textbf{Zhixiang He}$^{2}$ & \textbf{Zhongjiang He}$^{2\dagger}$ \\
\multicolumn{2}{c}{\textbf{Kongming Liang}$^{1\ddagger}$} & \multicolumn{2}{c}{\textbf{Zhanyu Ma}$^{1}$}
\end{tabular}\par}
\vspace{0.75em}
{\normalsize $^{1}$School of Artificial Intelligence, Beijing University of Posts and Telecommunications\par}
\vspace{0.12em}
{\normalsize $^{2}$Institute of Artificial Intelligence (TeleAI), China Telecom\par}
\vspace{0.35em}
{\small
\texttt{\{2362589239,wuwenzhuo,jzz2023210187,yangyunhao\}@bupt.edu.cn}\par
\texttt{\{liangkongming,mazhanyu\}@bupt.edu.cn}\par
\texttt{\{zhanghy56,zangxh,hezx3\}@chinatelecom.cn}\par
\texttt{\{hezhongj1\}@163.com}\par}
\vspace{0.6em}
{\small Project Page: \url{https://kamisato520.github.io/DataEvolver}\par}
\end{center}
\begingroup
\renewcommand{\thefootnote}{\fnsymbol{footnote}}
\footnotetext[1]{Equal contribution.}
\footnotetext[2]{Project lead.}
\footnotetext[3]{Corresponding author.}
\endgroup
\setcounter{footnote}{0}
}

\begin{document}

\makecustomtitle

\begin{abstract}
Controllable visual data construction is becoming a bottleneck for image and video generation, editing, and multimodal understanding. In practice, useful supervision is rarely produced by a one-pass rendering script; it emerges from a closed-loop process of generation, inspection, correction, rerendering, filtering, and export. We refer to this perspective as \emph{Data-evolver}: useful data should be built by \emph{goal-driven loop agents} that repeatedly compare current artifacts against an explicit transformation goal, apply bounded corrections, and decide whether the result is fit to enter training. We instantiate this perspective with DataEvolver, a closed-loop multi-artifact visual data engine that makes the process explicit through persistent artifacts, bounded corrective actions, and explicit verdicts. The engine supports artifact-level records including RGB outputs, masks, depth maps, normal maps, meshes, poses, trajectories, transformation programs, and review traces.

In the current implementation, the goal-driven loop agents operate at two coupled levels: generation-time self-correction within each sample and validation-time self-expansion across dataset rounds. The current release validates only a minimal image-level object-rotation setting; other export modes are interface-supported but not yet benchmark-validated. Using a fixed \qwenedit{} LoRA probe, the final \textbf{Ours +DualGate} configuration outperforms the unadapted base model and a public multi-angle LoRA on both \spatialedit{} \cite{xiao2026spatialedit} and the held-out Eval1 test set. A four-stage ablation further reveals a clear construction pathway: \textbf{Ours-Base} provides a strong scene-aware starting point, \textbf{Ours +Feedback} improves semantic and viewpoint metrics, \textbf{Ours +InnerGate} further improves traditional image-quality metrics, and \textbf{Ours +DualGate} performs best overall on both traditional and VIE metrics. The broader contribution is therefore not only the rotation dataset itself, but a reusable Data-evolver framing in which visual data is built and improved through explicit goal tracking, review, correction, and acceptance loops.
\end{abstract}

\section{Introduction}
\label{sec:introduction}

Controllable visual data construction is becoming a bottleneck for image and video generation, editing, and multimodal understanding. Many tasks require not only RGB outputs, but also geometry, masks, poses, trajectories, and quality-control traces. However, most synthetic-data pipelines are still organized as one-pass rendering scripts, making it difficult to diagnose failures, reuse intermediate artifacts, or scale to new visual data formats.

In practice, visual data construction is rarely a single forward pass. Builders render assets, inspect outputs, patch prompts or scene parameters, regenerate samples, compare failures across rounds, and decide whether the result should be accepted, filtered, or rebuilt. This is a stateful closed loop rather than a loose script chain. Intermediate artifacts, review signals, and downstream validation all affect what the builder should do next, yet much of that operational knowledge usually remains implicit and difficult to reuse.

This report argues for a Data-evolver view of controllable data construction. The core claim behind the title is that useful data should be built and improved through \emph{goal-driven loop agents}: controllers that make the target contract explicit, review whether current artifacts satisfy that contract, select bounded corrections when they do not, and accumulate downstream feedback into later construction rounds. We instantiate this view with DataEvolver, a workflow-as-skill visual data engine for end-to-end automated controllable data construction. In this instantiation, VLM/CV feedback and AI coding agents carry out the same observation--diagnosis--adjustment cycle that human builders historically performed by hand, while the same controllable scene-state trajectory can be exported as either an image or a video record and retain geometry and review traces within the same sample. The broader scope of the engine is summarized in \cref{sec:capability}.

The current object-rotation image task is a minimal validation case, not the boundary of the Data-evolver perspective. We instantiate the framework on scene-aware object rotation images, where the object should rotate while the scene and camera remain fixed, because this offers a clean image-level transformation contract with clear geometric failure modes. Under a fixed downstream LoRA probe, the final \textbf{Ours +DualGate} configuration achieves the strongest performance against external baselines. The ablation chain further shows how this result is obtained: \textbf{Ours-Base} provides a strong scene-aware starting point, \textbf{Ours +Feedback} improves semantic and viewpoint metrics, \textbf{Ours +InnerGate} further improves traditional image-quality metrics, and \textbf{Ours +DualGate} is the best overall configuration on both traditional and VIE metrics. Concrete metrics are reported in \cref{sec:results}.

The contribution is fourfold:
\begin{itemize}
  \item We formulate Data-evolver as a goal-driven-loop-agent view of visual data construction and instantiate it as DataEvolver, a closed-loop visual data engine with explicit stages, review channels, bounded controller actions, persistent state, and verdict logic.
  \item We define a multi-artifact data schema and export interface for preserving images, videos, masks, depth, normals, meshes, poses, trajectories, transformation programs, and review traces within the same sample record.
  \item We introduce a dual-loop self-evolution mechanism in which an inner loop corrects generation-time failures and an outer loop uses downstream validation to target the next data-expansion step.
  \item We provide a minimal image-level case study on scene-aware object rotation and show a four-stage ablation chain in which outer feedback, inner quality gating, and external VLM post gating progressively improve downstream editing performance.
\end{itemize}

\section{Capability Overview for Goal-Driven Loop Agents}
\label{sec:capability}

DataEvolver is the current engine instantiation of the Data-evolver view. Under this view, a controllable scene state can be exported in multiple supervision formats rather than committed to a single image-pair interface, and goal-driven loop agents operate over that artifact space rather than over isolated final renders. The abstraction is broader than the current benchmarked release, so this section distinguishes clearly between what is validated now and what is already represented by the artifact schema and export interface.

\subsection{Supported Output Modalities}

The artifact schema is intentionally broader than a source--target RGB pair. A single construction round can preserve not only final renders but also geometry, temporal state, and quality-control metadata.

\begin{table}[t]
\centering
\footnotesize
\caption{Supported output modalities in the DataEvolver artifact schema.}
\label{tab:capability-modalities}
\begin{tabularx}{\linewidth}{@{} l l X @{}}
\toprule
Output modality & Current status & Example export record \\
\midrule
Static image pair & validated & Source RGB image, target RGB image, and instruction. \\
Multi-view image set & interface-supported & Canonical source view plus multiple target views from the same scene state. \\
Video sequence & interface-supported / planned benchmark & Time-indexed frames or a rendered clip with instruction and timestamps. \\
Per-frame mask & interface-supported & Binary or soft masks aligned to each frame or rendered view. \\
Per-frame depth & interface-supported & Depth map or depth sequence aligned to RGB outputs. \\
Per-frame normal & interface-supported & Surface normal map or normal sequence for geometry-aware supervision. \\
Object pose & interface-supported & Object transformation state for each exported view or frame. \\
Camera pose & interface-supported & Camera intrinsics/extrinsics for each rendered state. \\
Object trajectory & interface-supported & Time-indexed object path, motion curve, or transformation sequence. \\
Transformation script & validated at image level & Action program describing the requested controllable edit. \\
Review and verdict metadata & validated & Issue tags, scores, corrective actions, and acceptance state. \\
\bottomrule
\end{tabularx}
\end{table}

\subsection{Supported Transformation Families}

The same engine abstraction supports multiple controllable transformations. What changes is the action program and export mode, not the core construction paradigm.

\begin{table}[t]
\centering
\footnotesize
\caption{Representative transformation families and their image/video export modes.}
\label{tab:transformation-families}
\begin{tabularx}{\linewidth}{@{} l X X @{}}
\toprule
Transformation family & Image output & Video output \\
\midrule
Rotation & Before/after pair & Smooth rotation sequence \\
Translation & Source-target placement & Object moving trajectory \\
Scaling & Size-change pair & Zoom-in or zoom-out object motion \\
Camera motion & Multi-view frames & Camera trajectory video \\
Object insertion & Inserted image & Object appearing or being placed \\
Multi-object relation & Relation pair & Relation transition video \\
Compositional edit & Final edited image & Multi-step edit video \\
\bottomrule
\end{tabularx}
\end{table}

\subsection{Current Release Scope}

The capability overview above describes the broader Data-evolver abstraction, not the current experimental boundary. The current report validates only the image-level object-rotation setting. Other output types are supported by the rendering/export interface but are left as extension targets. This technical report focuses on establishing the engine abstraction and validating goal-driven loop agents on a simple image-level task, while leaving large-scale video and multi-task validation to future releases.
\section{Why Controllable Visual Data Requires Closed-Loop Construction}
\label{sec:motivation}

Controllable visual data should not be judged only by whether a file is rendered. A usable sample must satisfy the requested transformation while preserving the right invariants across RGB appearance, geometry, and, when present, time. Once the engine exports masks, depth, normals, poses, or trajectories in addition to RGB, the failure surface becomes broader rather than narrower: a single scene mistake can corrupt multiple artifacts at once. This is the generalized form of the original repeated Blender tuning problem.

\begin{table}[t]
\centering
\footnotesize
\caption{Typical failure modes for controllable visual data in image and video settings.}
\label{tab:failure-modes}
\begin{tabularx}{\linewidth}{@{} l X X @{}}
\toprule
Failure type & Image case & Video case \\
\midrule
Grounding failure & Object floats above or penetrates the support surface. & Object floats or penetrates the surface during motion. \\
Scale failure & Single-frame object scale is implausible. & Object size jitters across time or motion range. \\
Lighting failure & Render is overexposed, underexposed, or visually inconsistent. & Temporal flicker or lighting drift appears across frames. \\
Pose failure & Target angle or pose is incorrect. & Trajectory becomes discontinuous or semantically wrong. \\
Mask failure & The mask misses object regions or leaks into background. & Per-frame masks jump, flicker, or drift over time. \\
Geometry failure & Depth or normals are implausible for the rendered image. & Depth or normal sequences become temporally inconsistent. \\
Identity failure & Object identity changes within a single edited frame. & Object identity drifts across the video sequence. \\
Camera/object ambiguity & It is unclear whether the camera moved or the object changed. & Trajectory semantics become ambiguous across the sequence. \\
\bottomrule
\end{tabularx}
\end{table}

Multi-artifact export amplifies failure propagation: a single bad object pose may simultaneously corrupt the RGB target, mask alignment, depth map, normal map, trajectory label, and downstream training pair. As the engine preserves more artifacts, quality control becomes more important rather than less.

These failure modes explain why ad-hoc script chains are fragile. A one-pass pipeline can generate images, masks, meshes, renders, and training pairs, but it does not by itself encode why an artifact should be trusted. It also does not preserve enough state to explain downstream regressions. If a model trained on the resulting dataset improves viewpoint alignment but loses identity consistency or temporal stability, the system needs to know which dataset round, asset subset, review signal, or action program produced the conflict.

Closed-loop construction is therefore a necessity rather than an implementation preference. The engine must support observation, diagnosis, bounded correction, rerendering, filtering, and explicit acceptance decisions. That logic applies to both image and video exports. The current image-level rotation study isolates this requirement in its simplest form, while the same motivation extends naturally to temporal motion, multi-object relations, and compositional spatial edits.

\section{Related Work}
\label{sec:related}

\paragraph{Image and video data generation pipelines.}
Image-editing datasets and scene/video generation pipelines increasingly package controllable supervision, but they usually emphasize final samples rather than construction traces. Step1X-Edit, ImgEdit, and Pico-Banana-400K target large-scale instruction-following image editing, while ZeroScene, SPATIALGEN, and Lyra illustrate a parallel move toward controllable multi-view or sequence-aware scene generation \cite{liu2025step1xedit,ye2025imgedit,qian2025picobanana400k,tang2025zeroscene,fang2025spatialgen,bahmani2025lyra}. DataEvolver is complementary to these efforts: it focuses on the construction process itself, keeping intermediate artifacts, review traces, and explicit quality verdicts attached to each dataset record.

\paragraph{Editing backbones and benchmarks.}
Recent open editing systems and benchmarks have broadened both model capability and evaluation scope. General-purpose backbones and resources such as \qwenedit{}, Step1X-Edit, ImgEdit, and Pico-Banana-400K reflect the current trend toward stronger instruction-following editing with more realistic data coverage \cite{wu2025qwenimage,qwen2025qwenimageedit2511,liu2025step1xedit,ye2025imgedit,qian2025picobanana400k}. Benchmarks including \spatialedit{}, PICABench, and UniREditBench make fine-grained spatial control, physical realism, and reasoning-heavy edits more measurable beyond coarse semantic success \cite{xiao2026spatialedit,pu2025picabench,han2025unireditbench}. Our work is upstream of that stack: it asks how train-ready supervision is constructed before those backbones and benchmarks are applied.

\paragraph{Geometry-rich supervision and controllable \texorpdfstring{3D}{3D} assets.}
Promptable segmentation, image-to-3D reconstruction, and controllable asset construction make it increasingly practical to preserve geometry-rich intermediate artifacts during dataset building. SAM 3, Hunyuan3D 2.1, Hunyuan3D 2.5, Step1X-3D, Seed3D 1.0, PhysX, and Hunyuan3D Studio provide relevant background for segmentation, mesh recovery, material generation, and simulation-oriented asset preparation \cite{carion2025sam3segmentconcepts,teamhunyuan3d2025hunyuan3d21,lai2025hunyuan3d25,li2025step1x3d,feng2025seed3d,cao2025physx,lei2025hunyuan3dstudio}. Related work on high-fidelity texture and geometry transfer further motivates treating masks, depth, normals, and meshes as first-class artifacts rather than disposable implementation details \cite{yang2025highfidelity25dlatents,rampini2025texturemapping}.

\paragraph{Object-level control and structured editing logic.}
A parallel line of work improves controllability inside the editor itself through program decomposition, correspondence-aware editing, precise object-level control, part-level control, multimodal chain reasoning, interactive action spaces, causal object removal, and unified multimodal masked diffusion \cite{hu2025imageeditingprograms,almohammadi2025cora,schouten2025poem,cvejic2025partedit,zou2025beyondtextualcot,yu2026i2e,zhu2025georemover,li2025lavidao}. These works motivate our use of structured review signals and bounded actions, but our contribution is different: we do not propose a new editing backbone or an in-model planner. We instead define an engine abstraction for how controllable supervision is produced, inspected, corrected, and accepted.

\paragraph{Scene-aware and self-correcting construction.}
Scene-consistent editing and generation work highlights the importance of viewpoint stability, geometry, and explicit scene state \cite{baron2025editp23,tang2025zeroscene,fang2025spatialgen,bahmani2025lyra}. DataEvolver builds on that intuition but shifts the emphasis from final edited outputs to the construction loop that creates them. In the language of this paper, Data-evolver is goal-centric rather than sample-centric: goal-driven loop agents care about whether the construction trace is moving artifacts toward an explicit target contract. In contrast to sample-centric dataset releases, DataEvolver treats the construction trace, corrective action, and acceptance verdict as part of the dataset record.

\section{Goal-Driven Loop-Agent Formulation}
\label{sec:workflow}

DataEvolver operationalizes the Data-evolver view as a reusable, stateful, and inspectable procedure that turns a visual data request into staged artifacts, review signals, controller actions, and explicit verdicts. The key claim is that data construction should be governed by goal-driven loop agents rather than by a one-shot render call: the system must make the target transformation contract explicit, evaluate whether the current state satisfies that goal, and choose bounded next actions accordingly. At the workflow level, the skill is summarized as:

\begin{equation}
\label{eq:skill}
\skilltuple.
\end{equation}

\emph{Stages} are the ordered construction modules and the artifacts they must produce. \emph{Controller} is the AI agent or workflow manager that tracks state, routes samples, edits scripts, adjusts parameters, and decides which stage should run next. \emph{Review} contains automatic CV signals, VLM assessments, human-readable traces, geometry checks, and downstream validation metrics. \emph{Verdict} maps the observed evidence into actions such as \continueverdict{}, \inspect{}, \regenerate{}, \rejectverdict{}, \stoporrevert{}, or \nosignal{}.

This tuple explains how the workflow behaves, but it does not yet specify what a sample contains. To make the sample record explicit, DataEvolver organizes each request as an artifact graph:

\begin{equation}
\label{eq:artifact-graph}
\mathcal{G} =
(\mathcal{S}_{\mathrm{scene}}, \mathcal{A}_{\mathrm{asset}}, \mathcal{P}_{\mathrm{action}},
\mathcal{R}_{\mathrm{render}}, \mathcal{M}_{\mathrm{geom}},
\mathcal{T}_{\mathrm{temporal}}, \mathcal{Q}_{\mathrm{review}},
\mathcal{V}_{\mathrm{verdict}}, \mathcal{E}_{\mathrm{export}}).
\end{equation}

\(\mathcal{S}_{\mathrm{scene}}\) stores environment, lighting, support surfaces, and camera rigs. \(\mathcal{A}_{\mathrm{asset}}\) stores objects, meshes, materials, and semantic identifiers. \(\mathcal{P}_{\mathrm{action}}\) stores the transformation program, such as rotation, translation, scaling, insertion, or camera motion. \(\mathcal{R}_{\mathrm{render}}\) stores image or video outputs. \(\mathcal{M}_{\mathrm{geom}}\) stores masks, depth, normals, and pose-aligned records. \(\mathcal{T}_{\mathrm{temporal}}\) stores frame indices, trajectories, and motion curves. For static image pairs, this temporal state degenerates to two sampled states; for videos, it stores dense frame-wise state trajectories. \(\mathcal{Q}_{\mathrm{review}}\) stores VLM/CV judgments and diagnostic traces. \(\mathcal{V}_{\mathrm{verdict}}\) stores acceptance decisions. \(\mathcal{E}_{\mathrm{export}}\) stores the final output form, such as an image pair, a sequence, a trajectory dataset, preference data, or diagnostic logs.

\begin{table}[t]
\centering
\footnotesize
\caption{Operational contract of the goal-driven-loop-agent data engine.}
\label{tab:operational-contract}
\begin{tabularx}{\linewidth}{@{} l X @{}}
\toprule
Component & Contract \\
\midrule
Stage outputs & Construction modules emit scene states, assets, action programs, RGB renders, videos, masks, depth maps, normals, poses, trajectories, review traces, train-ready exports, checkpoints, and comparison reports that can be inspected or reused. \\
Artifact graph & Every sample preserves scene, asset, action, geometry, temporal, review, verdict, and export records rather than only a final RGB file. \\
Controller actions & The controller routes samples, adjusts parameters, edits workflow code when needed, triggers rerendering or regeneration, and manages state across rounds. \\
Review signals & Quality is assessed through CV signals, VLM feedback during construction, geometry checks, temporal diagnostics, and downstream benchmark behavior after training. \\
Feedback bounds & Updates remain bounded to actions such as rerendering, regenerating, filtering, rejecting, or delaying acceptance. \\
Verdict logic & A pair, sequence, or dataset round is accepted only if useful improvements do not trigger regression guards; mixed evidence is routed to \inspect{}. \\
\bottomrule
\end{tabularx}
\end{table}

Static and temporal exports are therefore specializations of the same artifact graph. The value of this abstraction is traceability. A data engine should not only produce examples; it should explain which goal each example was trying to satisfy, why it was judged good enough to enter training, and what corrections were attempted before acceptance. Persistent state makes failures traceable, review signals make quality visible, and verdicts make iteration decisions explicit. Downstream training remains important, but it functions as a validation probe for goal-driven loop agents rather than the main method.

\begin{figure}[!htbp]
\centering
\includegraphics[width=0.80\linewidth]{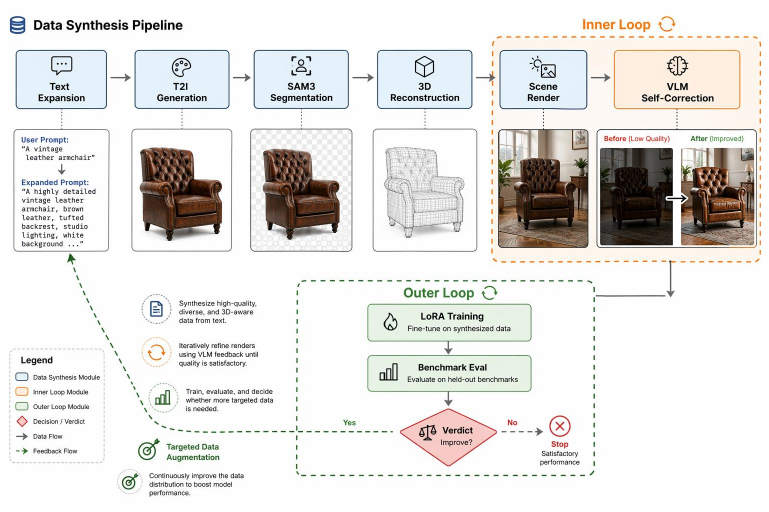}
\caption{Overall goal-driven-loop-agent workflow engine. A visual data request is converted into an artifact graph, reviewed by VLM/CV signals, routed through explicit verdicts, and validated by downstream probes rather than silently merged into training data.}
\label{fig:workflow-skill}
\end{figure}

\section{Dual-Loop Self-Evolving Visual Data Construction}
\label{sec:dual-loop}

DataEvolver contains two connected loops that together realize the behavior of goal-driven loop agents. The inner loop operates during generation and correction of individual samples or sequences. The outer loop operates after export, downstream training, and validation. Both loops share the same structure: review, bounded action, and verdict. They differ mainly in time scale and in what evidence they consume. At a high level, this decomposition is related to recent editing systems that make intermediate control steps explicit through programs, object-level control, or multimodal interaction, but our loops operate over data construction rather than single-sample inference \cite{hu2025imageeditingprograms,schouten2025poem,zou2025beyondtextualcot,yu2026i2e}.

The inner loop is generation-time self-correction. A candidate render or sequence is reviewed for lighting, grounding, viewpoint, scale, object completeness, material plausibility, mask integrity, and background consistency. The controller then adjusts Blender parameters, camera or framing, asset placement, materials, or filtering rules, and either rerenders, regenerates, or rejects the sample. Using a multimodal reviewer in this way is also consistent with recent benchmark practice, where spatial or physical editing quality is increasingly judged with VLM-assisted protocols rather than only pixel metrics \cite{xiao2026spatialedit,pu2025picabench,han2025unireditbench}.

For video construction, the inner loop additionally reviews temporal continuity, object trajectory smoothness, mask stability, depth consistency, and flicker. The outer loop can identify weak motion types, weak temporal segments, or weak transformation ranges, then trigger targeted video sequence generation or filtering. The common pattern in both loops is goal pursuit: the current state is measured against a target contract, and the next action is chosen to reduce the gap.

\begin{table}[t]
\centering
\footnotesize
\caption{Image- and video-level roles of the dual-loop design.}
\label{tab:dual-loop}
\begin{tabularx}{\linewidth}{@{} l X X X X @{}}
\toprule
Loop & Image-level review & Video-level review & Action & Verdict \\
\midrule
Inner loop & Lighting, grounding, scale, pose, and mask quality. & Flicker, trajectory smoothness, per-frame mask stability, and depth continuity. & Rerender, rescale, adjust camera, fix keyframes, smooth trajectory, or regenerate sequence. & Accept image pair, accept video sequence, or reject unstable segment. \\
Outer loop & Weak angle, weak category, or weak prompt subset. & Weak motion type, weak time segment, or weak trajectory pattern. & Targeted resampling, filtering, gate updates, or export-rule changes. & Continue, inspect, regenerate, reject, or stop/revert a round. \\
\bottomrule
\end{tabularx}
\end{table}

\begin{figure}[!tbp]
\centering
\begin{minipage}[t]{0.465\linewidth}
\centering
\includegraphics[width=\linewidth]{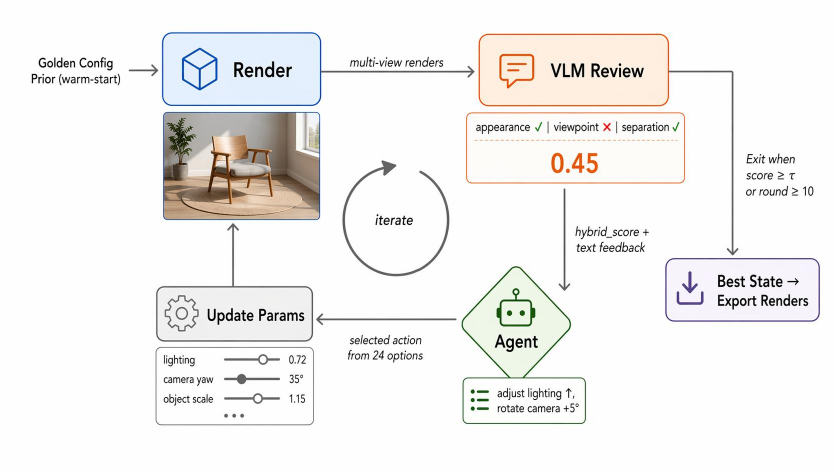}
\par\vspace{0.15em}
\small (a) Generation-time self-correction at the sample or sequence level.
\end{minipage}
\hfill
\begin{minipage}[t]{0.465\linewidth}
\centering
\includegraphics[width=\linewidth]{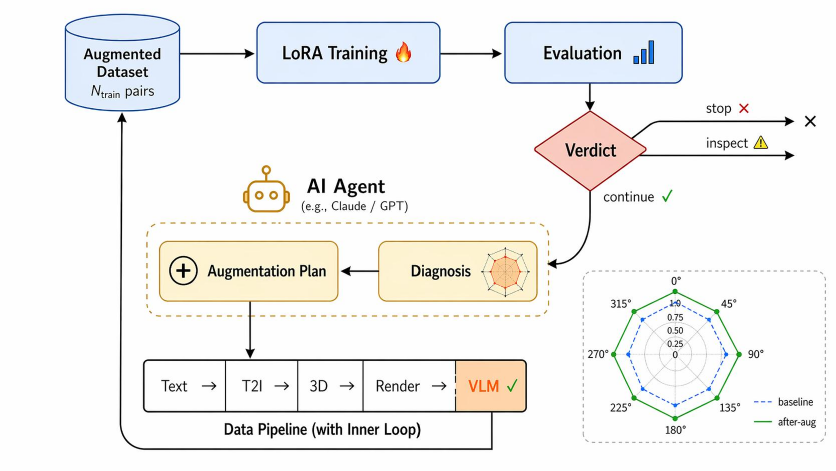}
\par\vspace{0.15em}
\small (b) Validation-time self-improvement at the dataset-round level.
\end{minipage}
\caption{Dual-loop self-evolving visual data construction under goal-driven loop agents. The inner loop corrects generation-time artifacts, while the outer loop converts downstream evaluation into the next data-construction decision.}
\label{fig:dual-loop}
\end{figure}

At the paper level, this system should be readable as a bounded state machine rather than as an informal agent capability. Review signals, bounded actions, safety checks, and stopping logic are part of the method definition. The appendix supplement records this operational contract in compact form, including the review fields, action-selection logic, safety conditions, stop criteria, and persisted logs needed to make the generation-time loop falsifiable (\cref{tab:inner-loop-contract}).

The two-loop design also explains why \inspect{} is a productive outcome. A mixed result is not merely a failed experiment; it is a signal that the outer loop found a useful direction but the inner loop did not yet enforce sufficient quality. Extending from images to videos does not require a different construction paradigm, but it does require additional temporal review signals, sequence-level verdict logic, and export rules for frame-wise artifacts and trajectories.

\section{Data-Build Pipeline Instantiation}
\label{sec:pipeline}

The current implementation instantiates DataEvolver as an export-mode pipeline for the broader Data-evolver view: a visual data request is converted into assets, a controlled scene state, an action program, a multi-artifact render bundle, and finally a chosen export format. The key point is that the engine does not start from an image pair; it starts from a controllable scene-state graph that can later be evaluated and revised by goal-driven loop agents before export as an image pair, a multi-view set, a video sequence, or a geometry package.

\subsection{Asset Preparation}

The pipeline begins with object concept expansion because the engine needs explicit object identities, descriptions, and synthesis prompts before visual generation can be controlled. White-background object generation or imported assets then provide object-centric visual seeds that are easier to segment and reconstruct than cluttered scenes. Foreground segmentation converts each object image into an RGBA asset suitable for image-to-3D reconstruction. This stage is aligned with promptable segmentation methods such as SAM 3 \cite{carion2025sam3segmentconcepts}, while the reconstruction stage is informed by recent asset-generation systems such as Hunyuan3D 2.1, Hunyuan3D 2.5, Step1X-3D, Seed3D 1.0, and PhysX \cite{teamhunyuan3d2025hunyuan3d21,lai2025hunyuan3d25,li2025step1x3d,feng2025seed3d,cao2025physx}. In the goal-driven-loop-agent view, these are artifact-producing stages rather than isolated scripts.

\subsection{Scene Setup}

Scene setup establishes the reusable environment into which assets are inserted. This stage configures scene templates, lighting, support planes, camera rigs, collision behavior, and scale normalization inside Blender \cite{blenderFoundationBlender}. It is also where grounding and framing failures become visible. Because the same scene state may later be exported as either images or videos, the setup must preserve stable camera semantics and consistent physical support rather than overfitting to a single frame.

\subsection{Action Program}

Each data sample is defined by an action program rather than by a direct render call. Representative action primitives are listed in \cref{tab:action-primitives}.

\begin{table}[t]
\centering
\footnotesize
\caption{Representative action primitives in the export-mode pipeline.}
\label{tab:action-primitives}
\begin{tabularx}{\linewidth}{@{} l X @{}}
\toprule
Action primitive & Example \\
\midrule
Rotation & \texttt{rotate(object, yaw=90)} \\
Translation & \texttt{translate(object, dx, dy, dz)} \\
Scaling & \texttt{scale(object, factor=1.2)} \\
Camera motion & \texttt{move\_camera(path=\dots)} \\
Composition & \texttt{compose([rotate, translate])} \\
\bottomrule
\end{tabularx}
\end{table}

In image mode, the engine may sample only the source and target states from this program. In video mode, the same program is sampled densely over time so that a trajectory such as yaw000 \(\rightarrow\) yaw015 \(\rightarrow\) yaw030 \(\rightarrow\) \dots \(\rightarrow\) yaw090 becomes a smooth motion sequence rather than a single pair. This keeps camera motion, object motion, and compositional edits explicit in the construction record.

\subsection{Multi-Artifact Rendering}

Given a scene state and action program, the renderer exports more than RGB. The render bundle can include RGB images or frames, masks or per-frame masks, depth maps or depth sequences, normal maps or normal sequences, semantic identifiers, object pose, camera pose, and transformation metadata. DataEvolver does not only render RGB outputs; it preserves geometry, temporal, and quality-control artifacts as first-class dataset records.

\subsection{Export Schema}

The final stage chooses the dataset export mode. Different downstream tasks require different views of the same artifact graph.

\begin{table}[t]
\centering
\footnotesize
\caption{Example export types produced by the same engine.}
\label{tab:export-schema}
\begin{tabularx}{\linewidth}{@{} l X @{}}
\toprule
Export type & Example \\
\midrule
Image pair & Source image + target image + instruction. \\
Image multi-view set & Canonical source + multiple target views from the same scene state. \\
Video sequence & Frame sequence + instruction + timestamps. \\
Video pair & Source video + edited video. \\
Geometry package & RGB + mask + depth + normal. \\
Trajectory dataset & Frames + object/camera pose over time. \\
Preference data & Accepted sample versus rejected sample or preferred sequence. \\
Diagnostic logs & Issue tags + actions + verdict records. \\
\bottomrule
\end{tabularx}
\end{table}

The current release uses the image-pair export mode because it maps cleanly to the downstream image-editing interface used by the validation probe, similar in spirit to object-level editing interfaces studied in work such as Cora, POEM, and PartEdit \cite{almohammadi2025cora,schouten2025poem,cvejic2025partedit}. The broader export logic, however, is not image-specific.

\section{Case Study I: Scene-Aware Object Rotation Images}
\label{sec:case-study}

This case study is intentionally simple. It validates the goal-driven-loop-agent principle under a controlled image-level transformation before scaling the same engine to video and richer geometry supervision. The current object-rotation image task is a minimal validation case, not the boundary of the broader Data-evolver perspective.

Rotation is chosen not because the engine is limited to rotation, but because it exposes viewpoint, grounding, identity, and scene-invariance failures while keeping the action program simple enough for a controlled ablation.

We use scene-aware object rotation because it preserves a clear change/invariance contract: the object viewpoint should change while the scene and camera remain fixed. This is a useful stress test for viewpoint accuracy, object identity preservation, grounding, and appearance consistency, and it also avoids confusing object motion with camera-orbit multiview rendering.

The task-specific rendering contract is:

\contractbox{\texttt{canonical yaw000 state} \(\rightarrow\) \texttt{rotate object yaw} \(\rightarrow\) \texttt{keep scene and camera fixed}}

By fixing the scene and camera and rotating only object yaw, the image pair directly encodes object-level rotation editing. In a video export mode, the same contract becomes a temporally sampled yaw trajectory rather than a single source-target pair: image mode uses yaw000 \(\rightarrow\) yaw090, whereas video mode can use yaw000 \(\rightarrow\) yaw015 \(\rightarrow\) yaw030 \(\rightarrow\) \dots \(\rightarrow\) yaw090.

The current release uses the canonical front view as the source and non-zero horizontal viewpoints as targets. Each training object contributes seven non-front target views, producing 245 training pairs from 35 training objects. The front-equivalent full-rotation slot is excluded from training because the canonical front view already serves as the source state and is reserved as an evaluation slot. Validation and test partitions remain object-disjoint, with 49 validation pairs and 56 test pairs.

The closed-loop ablation chain then adds three progressively stronger controls on top of this base export mode: an outer weak-angle feedback loop, an internal quality gate, and an external VLM post gate. The case study therefore connects the broader goal-driven-loop-agent abstraction to a concrete, falsifiable image-level protocol while leaving video, temporal motion, multi-object relation, and compositional edits as immediate extensions.

\section{Validation Protocol}
\label{sec:validation}

Validation is reported at two levels: a current case-study validation that measures whether the image-level rotation data changes downstream editing behavior, and an engine-level validation template that defines how future image, video, and multi-artifact releases should be assessed.

\subsection{Current Case-Study Validation}

The current validation probe is intentionally narrow. We use low-rank adapter fine-tuning on \qwenedit{} as a downstream probe; the Qwen-Image technical report and \qwenedit{} model card document the underlying image-editing backbone used in this probe \cite{wu2025qwenimage,qwen2025qwenimageedit2511}. Our work does not claim a new adapter method. The adapter is used only as an instrument for measuring whether the data engine produces useful supervision. We compare against the unadapted base model and a public multi-angle LoRA released by fal \cite{fal2026qwenmultianglelora}, while treating these as external baselines rather than controlled same-recipe data ablations. For the external comparison, we report the final \textbf{Ours +DualGate} configuration as the complete current Data-evolver system. The intermediate configurations are then analyzed separately in the four-stage ablation chain. The main external benchmark is \spatialedit{} \cite{xiao2026spatialedit}.

Before adding closed-loop scaling, the rotation probe went through a sequence of non-feedback ablations. These runs are named in the text by their design role, with the original internal experiment identifiers kept only to document implementation history.

\begin{table}[t]
\centering
\footnotesize
\caption{Ablation path from early rotation probes to the reader-facing four-stage chain used in the results.}
\label{tab:experiment-evolution}
\begin{tabularx}{\linewidth}{@{} l l >{\raggedright\arraybackslash}X @{}}
\toprule
Internal ID & Reader-facing name & Main intervention and role \\
\midrule
exp1 & RGB Rotation LoRA & Initial feasibility probe using dark RGB renders, no explicit direction word, and no object localization. \\
exp3 & Target-Localized BBox LoRA & Adds visual target localization through bounding-box conditioning; useful for diagnosis, but not a clean final interface. \\
exp4 & \brightprobe{} & Fixes raw image inputs, brightens rendering, and adds clockwise direction wording; strongest early non-feedback quantitative baseline. \\
exp5 & Ours-Base & Replaces the visual bbox dependency with object-description prompts and serves as the starting point of the closed-loop ablation chain. \\
R1/R2/R3 & Ours +Feedback / Ours +InnerGate / Ours +DualGate & Adds the outer feedback loop, then the internal quality gate, and finally the external VLM post gate. \\
\bottomrule
\end{tabularx}
\end{table}

Across this chain, the downstream training probe is held fixed at rank 32, learning rate \(1\times10^{-4}\), 30 training epochs, and the epoch 29 checkpoint for comparison. The prompt format follows the train-ready instruction field, using view-language instructions that ask the model to rotate an object from the front view to a target side or back view. These settings are reported for reproducibility, but the main claim remains about data construction and controlled workflow ablation.

\subsection{Engine-Level Validation Template}

Beyond the current image benchmark, DataEvolver should be evaluated as a construction engine for goal-driven loop agents. The relevant questions are not only whether the downstream model improves, but also whether the engine reliably produces complete, correctable, and accept-worthy artifact bundles that move toward the intended task goal.

\begin{table}[t]
\centering
\footnotesize
\caption{Engine-level validation dimensions for current and future releases.}
\label{tab:engine-validation}
\begin{tabularx}{\linewidth}{@{} l X X @{}}
\toprule
Metric & Image release & Video release \\
\midrule
Render success rate & Fraction of requested images rendered successfully. & Fraction of requested sequences rendered successfully. \\
Artifact completeness & Whether RGB, mask, depth, and other requested artifacts are present. & Whether each frame preserves the required artifact bundle. \\
Correction rounds & Average number of correction rounds per accepted image. & Average number of correction rounds per accepted sequence. \\
Acceptance rate & Fraction of image pairs that pass quality gates. & Fraction of video sequences that pass sequence-level gates. \\
Geometry validity & Grounding, clipping, and pose validity. & Temporal grounding, penetration, and geometry continuity. \\
Temporal consistency & Weakly applicable or not applicable. & Flicker, trajectory smoothness, and sequence consistency. \\
Review reliability & Agreement between VLM and CV signals on image artifacts. & Agreement between temporal VLM/CV signals on sequence artifacts. \\
\bottomrule
\end{tabularx}
\end{table}

\subsection{Future Video Validation}

The present report does not claim to have completed a video benchmark. A future video release should make the evaluation target explicit along four complementary axes.

\paragraph{Sequence-level metrics.}
These metrics evaluate whole-sequence success, such as overall render success rate, edit completion rate, and trajectory accuracy.

\paragraph{Frame-level artifact metrics.}
These metrics evaluate whether each frame preserves the requested artifact bundle, including RGB fidelity, per-frame mask IoU stability, depth-map availability, and normal-map availability.

\paragraph{Action-label metrics.}
These metrics test whether the rendered sequence matches the intended action program, including motion correctness, action-label consistency, and object/camera role disambiguation.

\paragraph{Temporal consistency metrics.}
These metrics evaluate flicker, optical-flow consistency, trajectory smoothness, depth temporal smoothness, and VLM-based temporal coherence. Together, they extend the same engine abstraction to sequence exports rather than replacing it.

\section{Results on the Minimal Image-Level Rotation Case}
\label{sec:results}

This section intentionally reports results only for the current image-level object-rotation release. The goal is to validate the goal-driven-loop-agent construction principle under a controlled export mode, not to claim that the full image-video-geometry capability space has already been benchmarked. The broader engine scope is summarized in \cref{sec:capability,sec:roadmap}.

\subsection{External Comparison of the Final Data-evolver System}
\label{sec:external-comparison}

We compare three reader-facing entries in the external benchmark: \textbf{Base} (\qwenedit{}), \textbf{Public LoRA} (the external \texttt{fal/Qwen-Image-Edit-2511-Multiple-Angles-LoRA} baseline), and \textbf{Ours +DualGate}, the final complete configuration of the current Data-evolver instantiation. This comparison evaluates the final system-level effect of the goal-driven-loop-agent data engine. The intermediate variants are analyzed separately in the ablation study in \cref{sec:feedback-results}.

On \spatialedit{} (488 pairs), \textbf{Ours +DualGate} is best on all five reported metrics, reaching PSNR 16.79, SSIM 0.7395, LPIPS 0.2478, CLIP-I 0.9525, and DINO 0.8918. Relative to Public LoRA, \textbf{Ours +DualGate} improves PSNR by 1.03 dB, SSIM by 0.0850, LPIPS by 0.0965 in the lower-is-better direction, CLIP-I by 0.0625, and DINO by 0.0474. This result provides the main final-system evidence: the complete goal-driven-loop-agent pipeline instantiated here by DataEvolver yields stronger rotation-editing supervision than both the unadapted base model and the public multi-angle LoRA baseline.

\begin{table}[!htbp]
\centering
\footnotesize
\caption{External comparison on \spatialedit{} (488 pairs) using Base, Public LoRA, and the final Ours +DualGate configuration.}
\label{tab:external-spatialedit-comparison}
\resizebox{\linewidth}{!}{
\begin{tabular}{lrrrrr}
\toprule
Method & PSNR $\uparrow$ & SSIM $\uparrow$ & LPIPS $\downarrow$ & CLIP-I $\uparrow$ & DINO $\uparrow$ \\
\midrule
Base (Qwen-Image-Edit-2511) & 14.97 & 0.6455 & 0.3625 & 0.8780 & 0.7832 \\
Public LoRA & 15.76 & 0.6545 & 0.3443 & 0.8900 & 0.8444 \\
Ours +DualGate & \textbf{16.79} & \textbf{0.7395} & \textbf{0.2478} & \textbf{0.9525} & \textbf{0.8918} \\
\bottomrule
\end{tabular}
}
\end{table}

\begin{figure}[!htbp]
\centering
\includegraphics[width=0.97\linewidth]{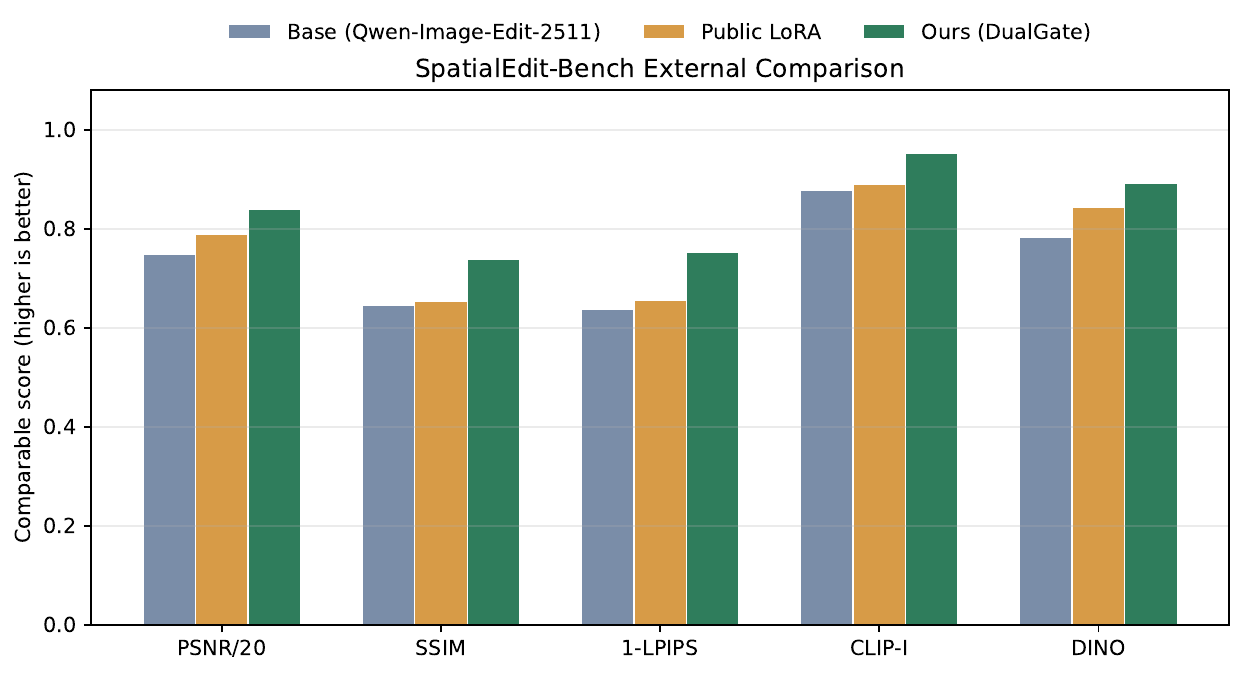}
\caption{External comparison on \spatialedit{}. The final Ours +DualGate system is best on PSNR, SSIM, LPIPS, CLIP-I, and DINO.}
\label{fig:external-spatialedit-metrics}
\end{figure}

The same conclusion holds on the Eval1 Test Set (56 pairs). \textbf{Ours +DualGate} again leads Base and Public LoRA across all five metrics, reaching PSNR 28.95, SSIM 0.9176, LPIPS 0.0855, CLIP-I 0.9640, and DINO 0.9282. Relative to Public LoRA, the gains are +13.75 dB PSNR, +0.1626 SSIM, -0.1895 LPIPS, +0.0340 CLIP-I, and +0.0332 DINO. The larger PSNR gain on Eval1 is expected because Eval1 is an in-domain held-out set constructed from the same rendering family, whereas \spatialedit{} is an external benchmark with broader distributional variation. Together with the \spatialedit{} comparison, this shows that the complete Data-evolver system yields a robust improvement across both the external benchmark and the held-out in-domain evaluation set.

\begin{table}[!htbp]
\centering
\footnotesize
\caption{External comparison on the Eval1 Test Set (56 pairs) using the same Base, Public LoRA, and final Ours +DualGate entries.}
\label{tab:external-testset-comparison}
\resizebox{\linewidth}{!}{
\begin{tabular}{lrrrrr}
\toprule
Method & PSNR $\uparrow$ & SSIM $\uparrow$ & LPIPS $\downarrow$ & CLIP-I $\uparrow$ & DINO $\uparrow$ \\
\midrule
Base (Qwen-Image-Edit-2511) & 14.92 & 0.7179 & 0.3145 & 0.9127 & 0.8787 \\
Public LoRA & 15.20 & 0.7550 & 0.2750 & 0.9300 & 0.8950 \\
Ours +DualGate & \textbf{28.95} & \textbf{0.9176} & \textbf{0.0855} & \textbf{0.9640} & \textbf{0.9282} \\
\bottomrule
\end{tabular}
}
\end{table}

\begin{figure}[!htbp]
\centering
\includegraphics[width=0.97\linewidth]{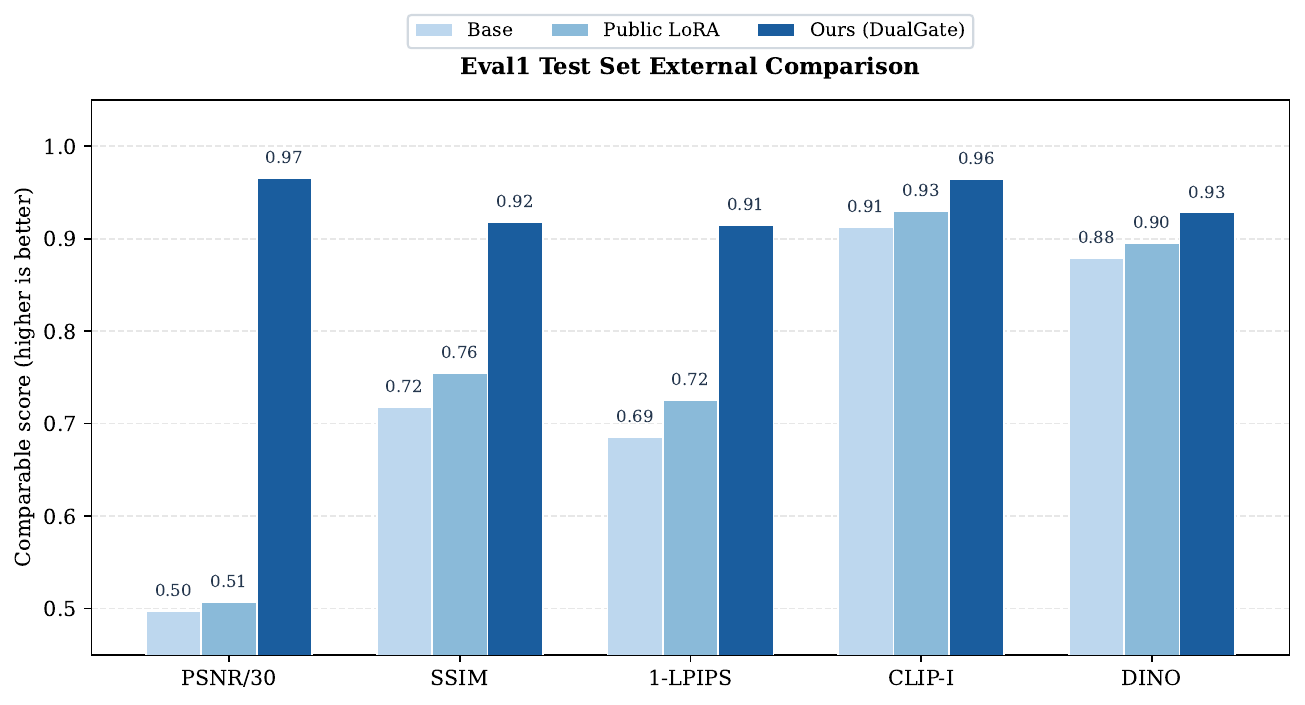}
\caption{External comparison on the Eval1 Test Set. The final Ours +DualGate system again leads Base and Public LoRA on all five metrics.}
\label{fig:external-testset-metrics}
\end{figure}

Appendix audit views further support the same conclusion. Direction-normalized summaries and per-angle breakdowns on both evaluation sets (\cref{fig:external-spatialedit-normalized-audit,fig:external-spatialedit-per-angle-psnr-audit,fig:external-spatialedit-per-angle-dino-audit,fig:external-testset-normalized-audit,fig:external-testset-per-angle-psnr-audit,fig:external-testset-per-angle-dino-audit}) show that the advantage of Ours +DualGate is not an artifact of a single aggregate metric.

\begin{figure}[!htbp]
\centering
\includegraphics[width=0.98\linewidth]{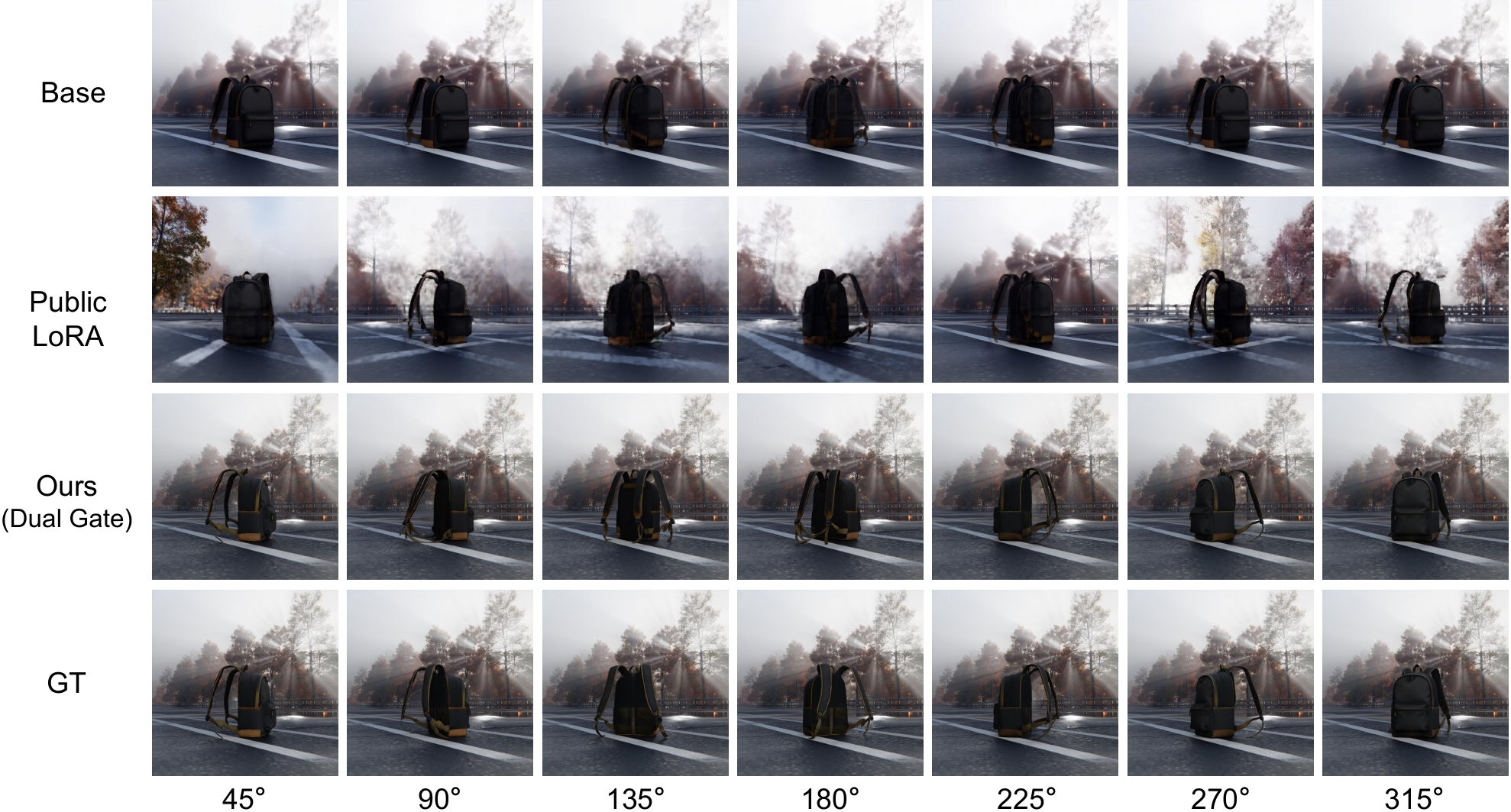}
\caption{Representative qualitative comparison on the held-out in-domain evaluation set constructed from the same rendering family as the training data. Ours +DualGate more closely follows the requested azimuth sequence and preserves object appearance better than Base and Public LoRA, both of which show visible artifacts in multiple views.}
\label{fig:qual-intra-domain}
\end{figure}

The quantitative gains are also visible in representative qualitative comparisons. On the held-out in-domain evaluation set shown in \cref{fig:qual-intra-domain}, Ours +DualGate tracks the target azimuth more accurately while better preserving object identity and overall visual quality across the rotation sequence. By contrast, both Base and Public LoRA show visible artifacts and weaker consistency in several views.

\begin{figure}[!htbp]
\centering
\includegraphics[width=0.94\linewidth]{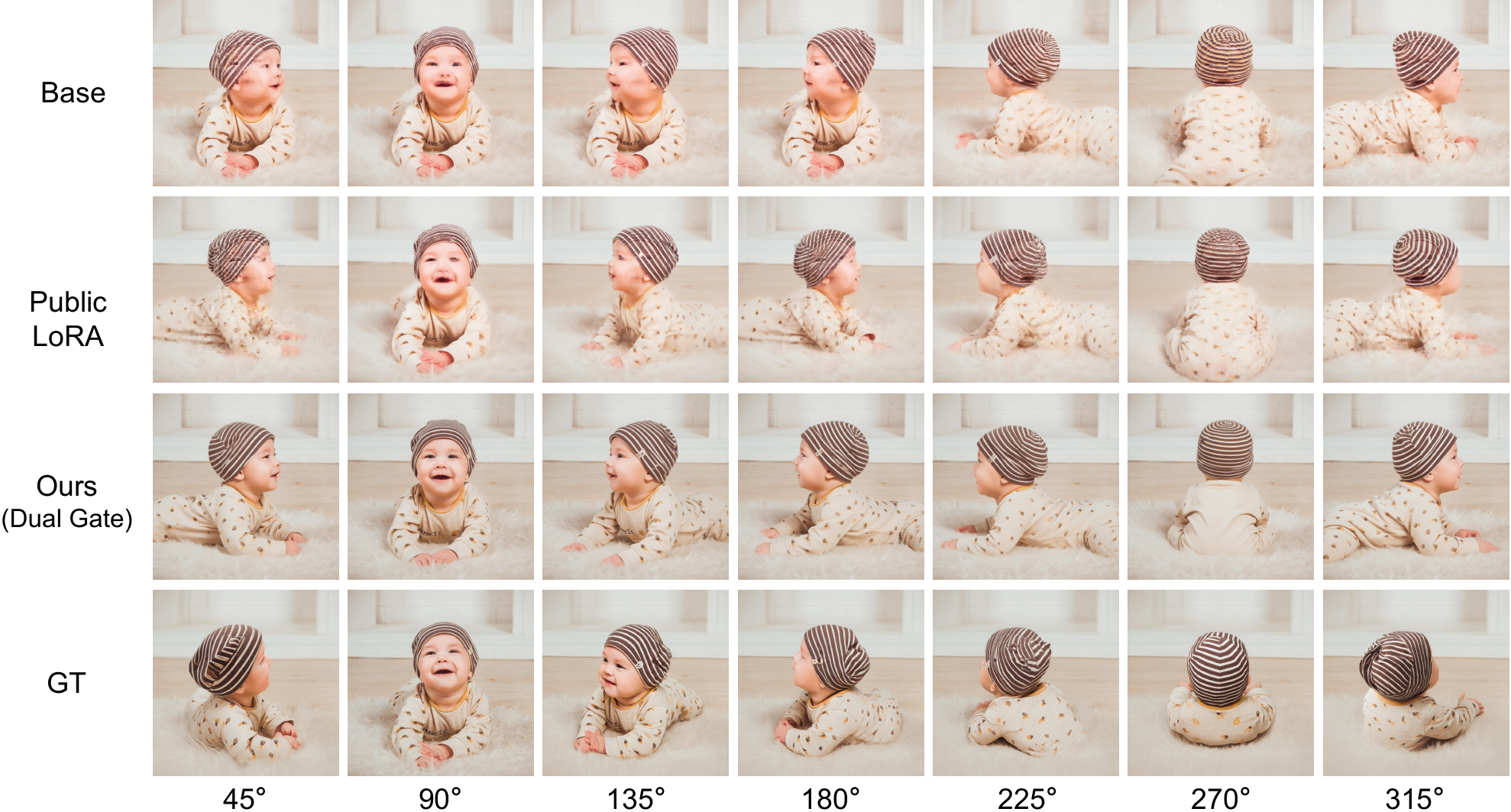}
\caption{Representative qualitative comparison on an out-of-domain example from the \spatialedit{} rotate subset. Ours +DualGate shows better viewpoint accuracy and object consistency than Base and Public LoRA, while the two external baselines exhibit visible artifacts and less stable structure.}
\label{fig:qual-extra-domain}
\end{figure}

\begin{figure}[!htbp]
\centering
\includegraphics[width=0.84\linewidth]{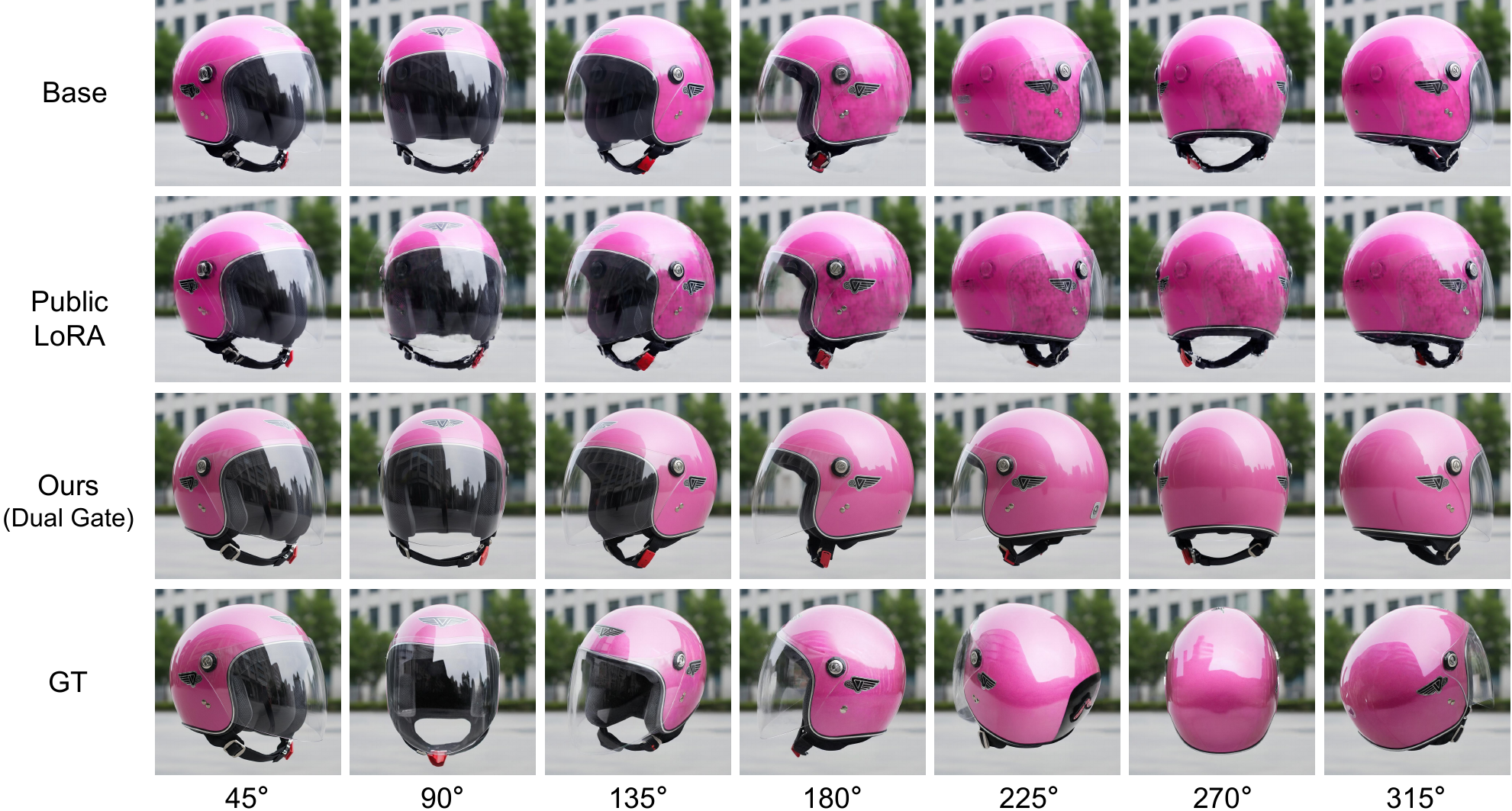}
\caption{A second out-of-domain qualitative comparison from the \spatialedit{} rotate subset. The same trend holds on a different object category: Ours +DualGate remains closer to the target rotation sequence and ground-truth appearance, whereas Base and Public LoRA again show strong artifact and consistency failures.}
\label{fig:qual-extra-domain-two}
\end{figure}

The same pattern persists on out-of-domain \spatialedit{} rotate examples (\cref{fig:qual-extra-domain,fig:qual-extra-domain-two}). Across two representative objects, Ours +DualGate more closely matches the requested viewpoint and preserves object structure more reliably. Base and Public LoRA again exhibit visible artifacts or unstable geometry, especially in side and rear views, which is consistent with the quantitative gap observed in both evaluation sets.

\subsection{Four-Stage Ablation Chain}
\label{sec:feedback-results}

The main ablation result is a monotonic four-stage improvement chain: \textbf{Ours-Base} (no feedback), \textbf{Ours +Feedback} (outer feedback loop only), \textbf{Ours +InnerGate} (outer feedback plus internal quality gate), and \textbf{Ours +DualGate} (outer feedback plus internal gate plus external VLM post gate). The chain is designed to isolate the contribution of each workflow component while keeping the downstream probe and evaluation protocol fixed.

\begin{table}[t]
\centering
\caption{Traditional ablation results on SpatialEdit-Bench (488 pairs).}
\label{tab:ablation_spatialedit}
\begin{tabular}{lccccc}
\toprule
Method & PSNR $\uparrow$ & SSIM $\uparrow$ & LPIPS $\downarrow$ & CLIP-I $\uparrow$ & DINO $\uparrow$ \\
\midrule
Ours-Base & 16.63 & 0.7296 & 0.2564 & 0.9050 & 0.8895 \\
Ours +Feedback & 16.68 & 0.7315 & 0.2538 & 0.9499 & 0.8903 \\
Ours +InnerGate & 16.74 & 0.7365 & 0.2502 & 0.9515 & 0.8912 \\
Ours +DualGate & 16.79 & 0.7395 & 0.2478 & 0.9525 & 0.8918 \\
\bottomrule
\end{tabular}
\end{table}

\begin{table}[t]
\centering
\caption{VIE Score ablation results on SpatialEdit-Bench (488 pairs). VIE Overall is computed as the arithmetic mean of Score\_view and Score\_cons.}
\label{tab:ablation_vie}
\begin{tabular}{lccc}
\toprule
Method & Score\_view $\uparrow$ & Score\_cons $\uparrow$ & VIE Overall $\uparrow$ \\
\midrule
Ours-Base & 0.7746 & 0.9682 & 0.8714 \\
Ours +Feedback & 0.7828 & 0.9685 & 0.8757 \\
Ours +InnerGate & 0.7852 & 0.9686 & 0.8769 \\
Ours +DualGate & 0.7888 & 0.9688 & 0.8788 \\
\bottomrule
\end{tabular}
\end{table}

\begin{figure}[!tbp]
\centering
\begin{subfigure}[t]{0.95\linewidth}
\centering
\includegraphics[width=\linewidth]{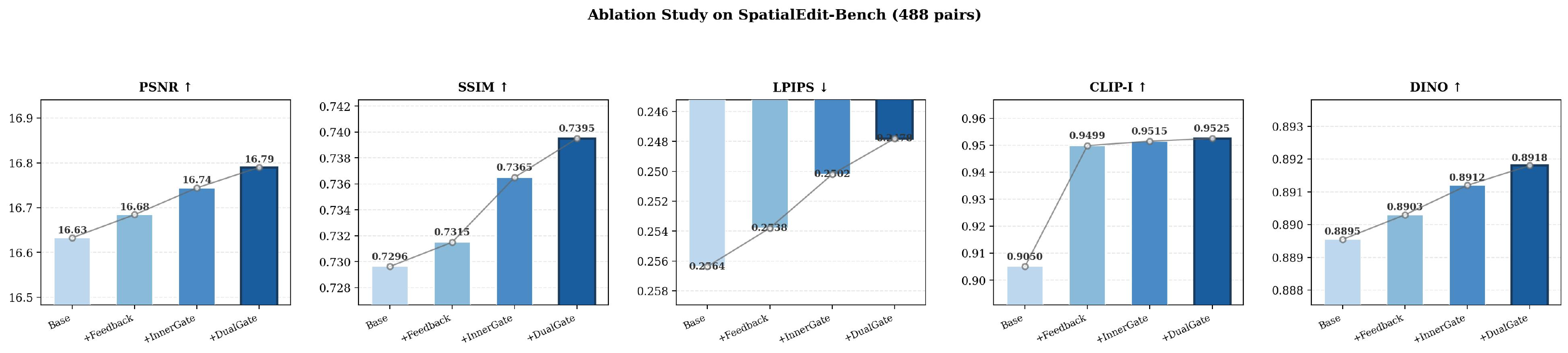}
\caption{Traditional metrics across the four-stage ablation chain.}
\label{fig:ablation-chain-metrics}
\end{subfigure}

\vspace{1.0em}

\begin{subfigure}[t]{0.90\linewidth}
\centering
\includegraphics[width=\linewidth]{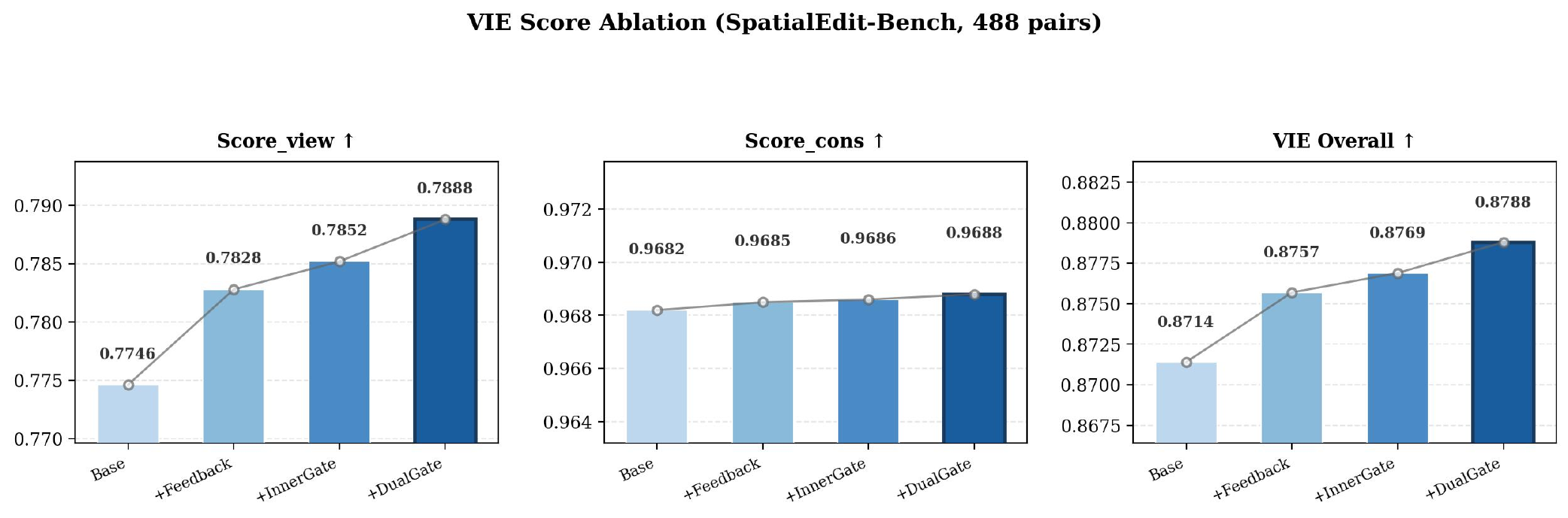}
\caption{VIE metrics across the same chain.}
\label{fig:ablation-chain-vie}
\end{subfigure}
\caption{Four-stage ablation chain for the goal-driven-loop-agent data engine. The final \textbf{Ours +DualGate} variant is best on both traditional and VIE metrics.}
\label{fig:ablation-chain-overall}
\end{figure}

\begin{figure}[!tbp]
\centering
\begin{subfigure}[t]{0.49\linewidth}
\centering
\includegraphics[width=\linewidth]{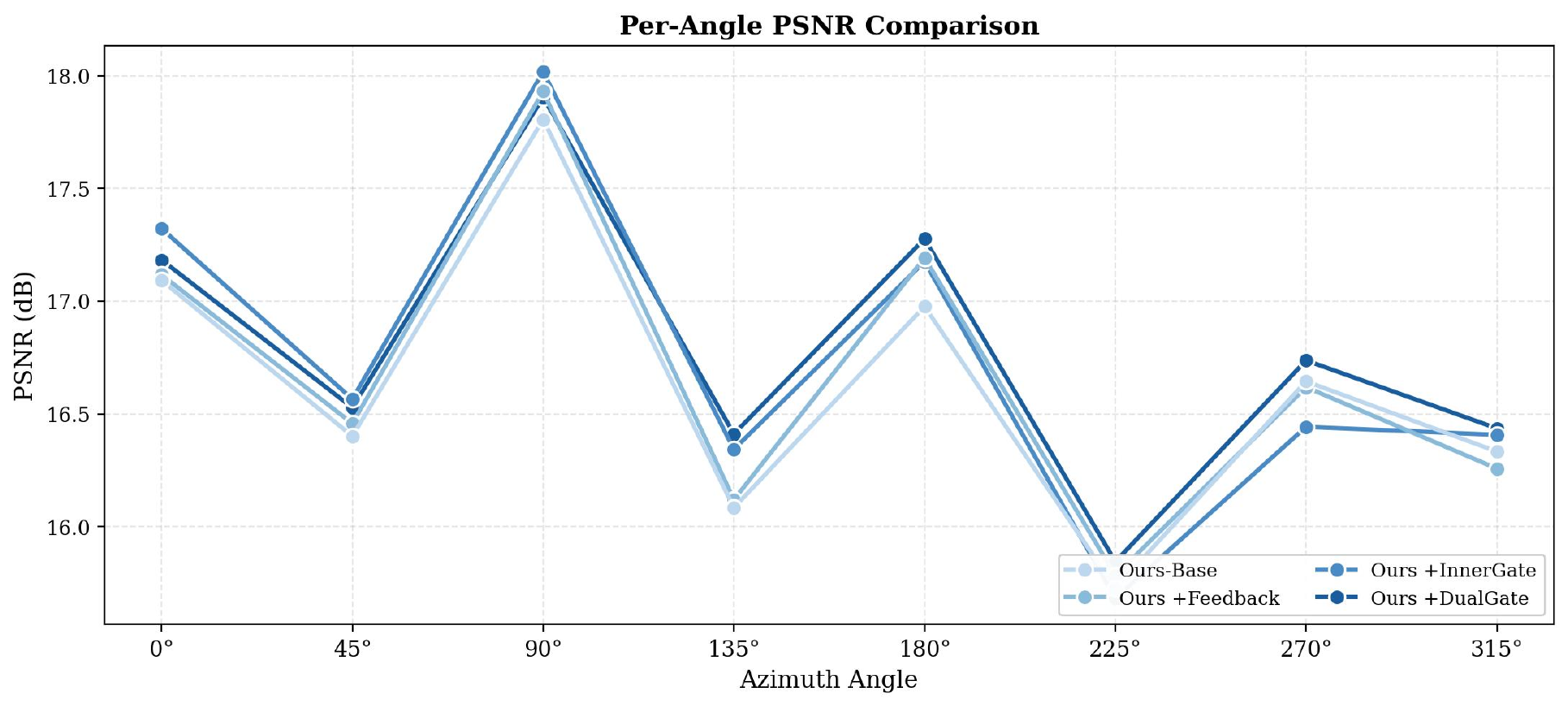}
\caption{Per-angle PSNR.}
\label{fig:ablation-chain-per-angle-psnr}
\end{subfigure}
\hfill
\begin{subfigure}[t]{0.49\linewidth}
\centering
\includegraphics[width=\linewidth]{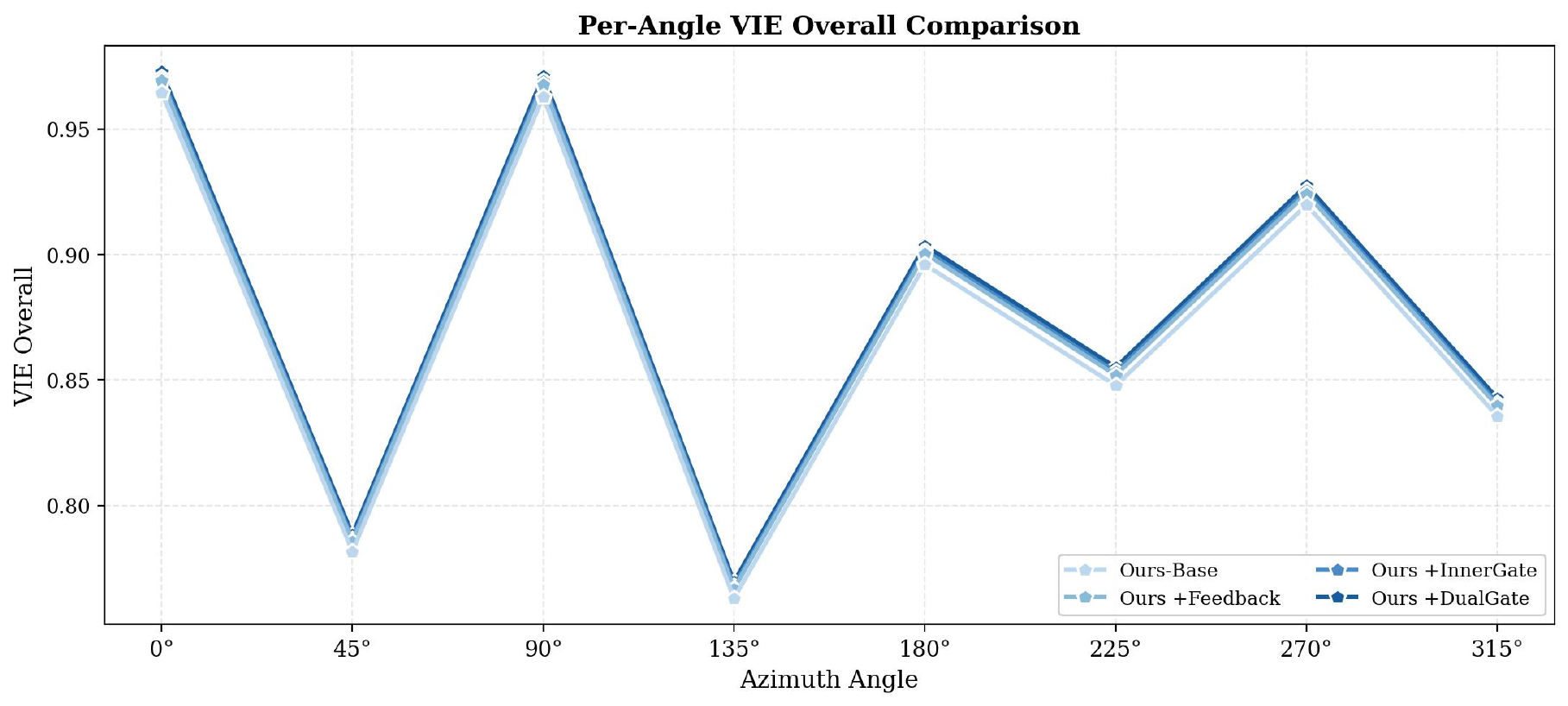}
\caption{Per-angle VIE Overall.}
\label{fig:ablation-chain-per-angle-vie}
\end{subfigure}
\caption{Per-angle diagnostics for the four-stage ablation chain. Improvements are visible in the overall averages and in several angle-specific subgroups, although the magnitude varies by angle. Because \spatialedit{} uses an eight-slot evaluation protocol, the per-angle diagnostic includes a front-equivalent full-rotation evaluation bin, whereas our constructed training split uses only the seven non-front target views. In the appendix audits, this front-equivalent evaluation slot is denoted as 360\(^\circ\).}
\label{fig:ablation-chain-per-angle}
\end{figure}

Several conclusions follow directly from Tables~\ref{tab:ablation_spatialedit} and~\ref{tab:ablation_vie} and Figures~\ref{fig:ablation-chain-overall} and~\ref{fig:ablation-chain-per-angle}. First, Ours-Base establishes a strong scene-aware starting point even before any closed-loop feedback is added. Although Table~\ref{tab:external-spatialedit-comparison} reports only the final Ours +DualGate system, the Ours-Base row in Table~\ref{tab:ablation_spatialedit} already exceeds the two external baselines in Table~\ref{tab:external-spatialedit-comparison} on all five traditional metrics. This shows that the base scene-aware data construction is already useful supervision. The remaining stages then explain how the complete Ours +DualGate system reported in the external comparison is obtained.

Second, adding only the outer feedback loop produces the largest semantic jump in the chain. Ours +Feedback improves CLIP-I from 0.9050 to 0.9499 and raises \scoreview{} from 0.7746 to 0.7828 relative to Ours-Base, while also improving PSNR, SSIM, LPIPS, \scorecons{}, and VIE Overall. This pattern shows that the external feedback loop successfully converts weak-angle evaluation signal into effective training-data gains rather than merely reshuffling prompts.

Third, the internal quality gate provides an additional quality-oriented refinement step. Moving from Ours +Feedback to Ours +InnerGate further improves PSNR, SSIM, and LPIPS, while also preserving the semantic gains in CLIP-I, DINO, and VIE. The most direct interpretation is that internal gating improves the quality of accepted training data, which then translates into cleaner pixel-level reconstruction and better overall consistency.

Finally, Ours +DualGate is the best configuration across both metric families. It achieves the strongest traditional results (PSNR 16.79, SSIM 0.7395, LPIPS 0.2478, CLIP-I 0.9525, DINO 0.8918) and also the strongest VIE results (\scoreview{} 0.7888, \scorecons{} 0.9688, VIE Overall 0.8788). This final step shows that combining the internal gate with the external VLM post gate completes the closed-loop data construction pipeline: feedback identifies where to scale, the inner gate filters generation quality, and the post gate consolidates the best train-ready additions.

The per-angle plots in \cref{fig:ablation-chain-per-angle} show that the gains are not confined to a single averaged summary. Both the PSNR and VIE curves improve across multiple angle subgroups, which indicates that the workflow is strengthening viewpoint control throughout the rotation range rather than overfitting to a narrow subset of views.

\begin{figure}[!htbp]
\centering
\includegraphics[width=0.96\linewidth]{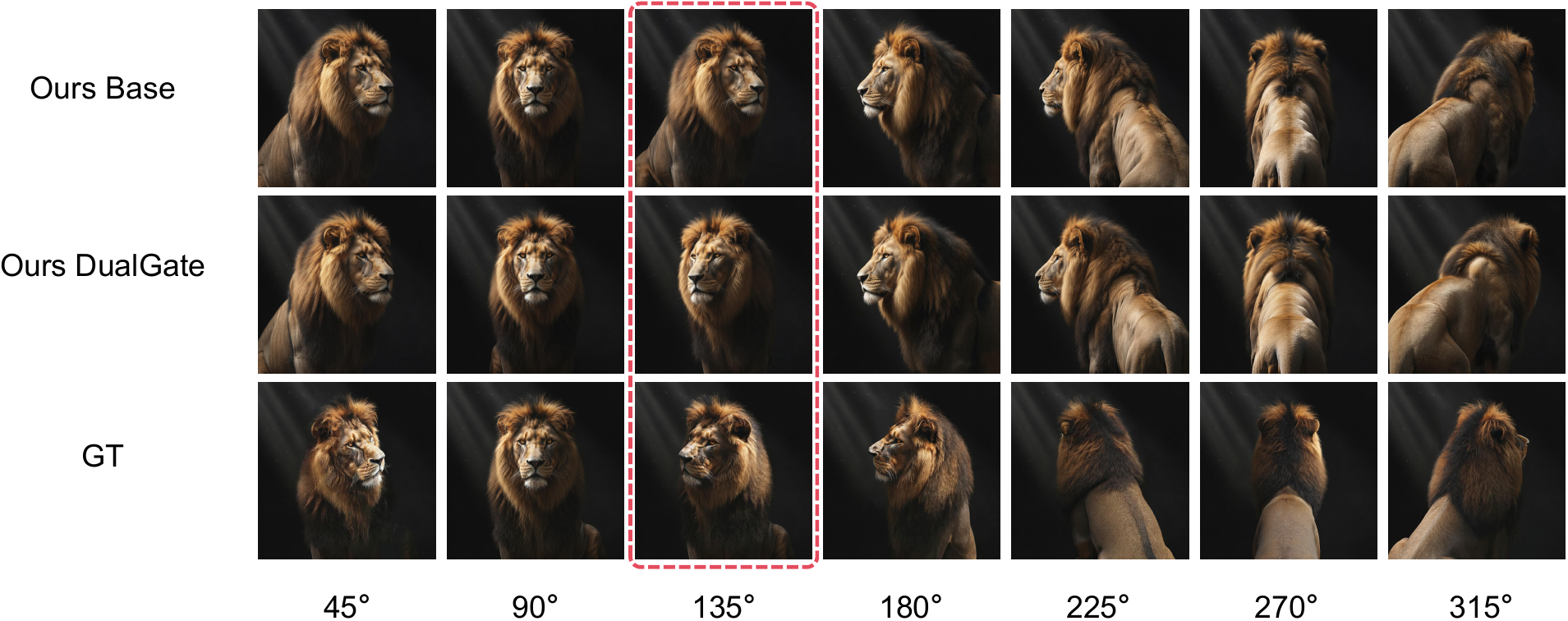}
\caption{Qualitative effect of the full closed-loop variant within the same model family. Compared with Ours-Base, Ours +DualGate produces a more coherent side/back transition and more stable object structure across neighboring viewpoints.}
\label{fig:qual-with-vs-without}
\end{figure}

The qualitative comparison in \cref{fig:qual-with-vs-without} illustrates the visual effect of adding the closed-loop data engine on top of the same model family. Relative to Ours-Base, the gated feedback variants produce smoother side-to-back transitions and more stable object structure across neighboring views.

Additional audit figures in the appendix provide direction-normalized changes and per-angle CLIP-I and DINO views of the same ablation chain (\cref{fig:ablation-normalized-audit,fig:ablation-clipi-audit,fig:ablation-dino-audit}). Together, these supplementary plots show that the improvement trend is consistent across pixel, semantic, and identity-sensitive metrics.

\subsection{Why the Minimal Case Matters}

This minimal case is informative because it isolates a controllable transformation with explicit invariants while still exercising nontrivial goal-driven loop agents. Scene-aware rotation requires the engine to coordinate asset quality, grounding, lighting, viewpoint language, and acceptance logic without hiding behind camera motion or scene changes. The monotonic ablation trend therefore validates the goal-driven-loop-agent mechanism itself: outer feedback identifies where new data should be added, inner gating improves what enters training, and post gating consolidates the strongest additions. What is validated here is the loop logic under a clean image contract, not the whole future roadmap.

\subsection{What This Case Study Validates and Does Not Validate}

Table~\ref{tab:validation-boundary} summarizes the boundary of the current release. The present study validates goal-driven-loop-agent construction under an image-level object-rotation contract, including object-disjoint data export, quality-gated supervision, downstream image-editing probing, and traceable artifact records. It does not yet validate long-horizon video editing, temporal gates, video-editing probes, or large-scale multi-task deployment. This distinction prevents over-claiming: the current evidence supports the engine logic, while future releases must test whether the same artifact graph and dual-loop design remain effective for sequence-level and multi-task data.

\begin{table}[t]
\centering
\footnotesize
\caption{Current validation boundary of the minimal image-level case study.}
\label{tab:validation-boundary}
\begin{tabularx}{\linewidth}{@{} X X @{}}
\toprule
Validated now & Not yet validated \\
\midrule
Closed-loop image data construction & Full video dataset training \\
Object rotation pair generation & Long-horizon temporal editing \\
Quality gates improve image supervision & Video temporal consistency gates \\
Downstream image-editing probe & Video-editing downstream probe \\
Artifact traceability & Large-scale multi-task deployment \\
\bottomrule
\end{tabularx}
\end{table}

\section{Extension Roadmap: From Image Rotation to Image-Video-Geometry Data Construction}
\label{sec:roadmap}

The present release validates the engine on a simple image-level task, but the Data-evolver abstraction is broader. Extending from images to videos mainly requires temporal review signals and sequence-level verdict logic, rather than a different construction paradigm. The roadmap below therefore follows the same artifact graph and dual-loop design while broadening the export modes of goal-driven loop agents.

\paragraph{Stage 1: Static geometry image data.}
The nearest extensions remain image-centric but already demonstrate the multi-artifact nature of the engine: rotation, translation, scaling, object placement, multi-view image sets, and joint RGB plus mask/depth/normal export examples.

\paragraph{Stage 2: Temporal visual data.}
The next step is sequence construction: smooth object rotation video, object translation video, object scaling video, camera trajectory video, object insertion processes, multi-object interaction clips, and before--after edited video pairs.

\paragraph{Stage 3: Compositional and relational data.}
Longer-term extensions combine actions and relations: move + rotate, rotate + scale, place object A behind object B, move A around B, model occlusion transitions, and export multi-step edit sequences or chain-of-edit supervision.

\begin{table}[t]
\centering
\footnotesize
\caption{Priority roadmap for future engine releases.}
\label{tab:roadmap-priority}
\begin{tabularx}{\linewidth}{@{} l l X @{}}
\toprule
Priority & Direction & Why \\
\midrule
P0 & Image + mask + depth + normal export examples & Easiest way to demonstrate multi-artifact capability beyond RGB-only pairs. \\
P1 & Smooth rotation video & Most natural temporal extension of the current rotation case. \\
P1 & Object translation video & Controllable, intuitive, and comparatively easy to evaluate. \\
P2 & Scale or placement video & Geometrically meaningful and compatible with existing scene-state controls. \\
P2 & Multi-object relation data & High paper value, but requires more complex relation-aware review logic. \\
P3 & Compositional video editing data & Highest long-term upside, but should follow after simpler temporal exports are stabilized. \\
\bottomrule
\end{tabularx}
\end{table}
\section{Discussion}
\label{sec:discussion}

The main value of DataEvolver is conceptual rather than task-specific. Under the Data-evolver framing, its contribution is an artifact-level construction paradigm in which scene state, action program, renderer, review signal, and export schema are defined explicitly and linked through verdict logic. Goal-driven loop agents then operate over that structured artifact space rather than over a single final render. Under this abstraction, images, videos, geometry maps, and temporal trajectories become different export views of the same controllable construction process.

This distinction matters because it separates paradigm from release scope. The current object-rotation study validates a narrow image-level contract, but the broader engine abstraction is intended to organize how future visual supervision is generated, corrected, and accepted. At the same time, generalization is not automatic: each new task still requires its own change/invariance contract, review signals, action space, and acceptance rules.

The practical lesson is that quality gates belong inside the data engine. If artifacts enter training merely because they were generated, scaling becomes unsafe. Render quality, geometry validity, temporal stability, and action-label consistency should therefore be treated as first-class acceptance conditions rather than after-the-fact diagnostics.

\section{Limitations}
\label{sec:limitations}

Current evidence is still image-centric. The only fully quantified study in this report is the scene-aware object-rotation image-pair setting, so the present release does not yet validate the full goal-driven-loop-agent story for video editing, temporal supervision, or large-scale multi-task deployment.

Video support is currently interface-level rather than benchmark-level. The engine can represent sequences, trajectories, and per-frame artifacts, but the report does not yet include downstream video-editing training or a standardized temporal benchmark.

Temporal quality control also remains less mature than image quality control. Video construction requires additional checks for flicker, trajectory smoothness, per-frame mask stability, depth continuity, and action-sequence consistency, and these checks are not yet as complete as the image-level gates used in the current case study.

Upstream asset quality is still a bottleneck across all export modes, and VLM review is not sufficient on its own: geometry validity, collision, grounding, pose accuracy, and temporal consistency still require programmatic checks. Real-data integration likewise remains future work.

\section{Conclusion}
\label{sec:conclusion}

This report presents Data-evolver as a view of controllable visual data construction and instantiates it with DataEvolver. Under this view, data construction is not a one-pass rendering pipeline but a closed-loop, multi-artifact process driven by goal-driven loop agents: the system generates, inspects, corrects, and exports images, videos, geometry maps, poses, trajectories, and review traces under explicit verdict logic, with inner-loop self-correction and outer-loop validation-time self-expansion.

The current scene-aware object-rotation study is intentionally a minimal image-level validation case. Even in this narrow setting, the final Ours +DualGate configuration outperforms the unadapted base model and a public multi-angle LoRA on both external and held-out evaluations. The four-stage ablation further explains this result: Ours-Base provides a strong scene-aware starting point, Ours +Feedback improves semantic and viewpoint behavior, Ours +InnerGate improves traditional image quality, and Ours +DualGate achieves the best overall performance on both traditional and VIE metrics.

The broader claim is therefore not that DataEvolver has already exhausted image, video, and geometry data construction. It is that controllable visual supervision should be built under a Data-evolver abstraction in which an inspectable artifact graph, bounded corrective actions, and explicit acceptance logic are all organized around goal-driven loop agents. That abstraction is what makes the current image case extensible to future video, temporal motion, multi-object relation, and compositional spatial-edit releases.

\appendix
\section{Reproducibility and Implementation Details}
\label{app:reproducibility}

The reproducible unit of the framework is a dataset-construction round. A round starts from a dataset goal and a set of object concepts. It then produces prompts, white-background object images, foreground masks, reconstructed meshes, scene-aware rendered views, review records, train-ready pairs, a downstream checkpoint, and a comparison report. Together, these artifacts form a linked state trace that lets later inspection connect a benchmark outcome to the samples and construction stages that produced it.

The case-study dataset uses a source-target pair schema. The source image is the canonical front view, and each target image corresponds to a requested horizontal rotation. Each row records the source image, target image, instruction text, target rotation, object identifier, object name, and prompt version. The instruction format uses explicit view names, for example asking the model to rotate a named object from the front view to a right-side or back view.

Split construction is object-disjoint. The Ours-Base configuration uses 35 train objects, 7 validation objects, and 8 test objects, yielding 245 train pairs, 49 validation pairs, and 56 test pairs. The subsequent gated-feedback variants keep the validation and test objects fixed while modifying the train-data construction workflow. This design prevents leakage of new synthetic objects into held-out splits and keeps the comparison focused on train-data interventions.

The training probe uses \qwenedit{} with low-rank adapter fine-tuning under a fixed comparison protocol \cite{wu2025qwenimage,qwen2025qwenimageedit2511}. The tracked runs use rank 32, learning rate \(1\times10^{-4}\), 30 epochs, and the epoch 29 checkpoint for evaluation. We evaluate the probe on \spatialedit{} \cite{xiao2026spatialedit}, which contains 488 rotation pairs across 61 objects and eight angle slots. The metric set combines traditional image metrics, representation metrics, distribution-level quality, and VLM-based view and consistency scores, matching the interpretation used in \cref{sec:results}.

Checkpoint and artifact availability remains partial in the current release package. The tracked comparison covers the reader-facing four-stage chain reported in the main text. Large checkpoint files and full inference image dumps are omitted for size reasons. The generated tables and figures bundled with this report provide the finalized benchmark views used in the paper.

A second supplement appendix (\cref{app:prism-reproducibility-supplement}) records additional implementation-level detail for the inner-loop contract and supplementary audit figures for the ablation chain.

\paragraph{Artifact index.}
Code, intermediate artifacts, and evaluation logs will be released in an anonymous repository upon publication. Key hyperparameters are LoRA rank \(=32\), learning rate \(=10^{-4}\), and training epoch \(=29\), held uniform across the compared rounds. Checkpoint hashes will be provided in the released repository.

\section{Supplement: Inner-Loop Operational Detail, External Comparison, and Ablation Audit}
\label{app:prism-reproducibility-supplement}

This supplement provides implementation-level detail that supports three claims in the main report. First, the generation-time inner loop can be described as a bounded state machine rather than as an informal agent capability. Second, the external-comparison claim in the main text is supported by supplementary audit figures that expose the same ordering from normalized, PSNR, and DINO viewpoints. Third, the four-stage ablation chain in the main text is supported by supplementary multi-metric audit figures that expose the same trend from normalized, semantic, and identity-consistency viewpoints.

\subsection{A Reproducible Inner-Loop State Machine}

The current implementation notes contain enough structure to describe the generation-time loop as a state machine rather than as an informal agent capability. Each review round receives a current render, optional previous render, optional original object reference, optional pseudo-reference scene insertion image, mask-derived metadata, and a bounded action space. The VLM review schema asks for five integer scores: lighting, object integrity, composition, render-quality semantics, and overall quality. It also asks for confidence values, issue tags, suggested actions, lighting diagnosis, structure consistency, color consistency, physics consistency, asset viability, and a pairwise comparison against the previous render when available. CV measurements provide exposure, sharpness, mask/framing terms when available, and programmatic physics checks such as contact gap and penetration. The hybrid score is computed as a weighted combination of VLM and CV signals; in scene-insertion mode the implementation note uses a 0.70 VLM weight and a 0.30 CV weight, aggregates per-view scores as \(0.7\times\mathrm{mean}+0.3\times\mathrm{worst}\), and applies hard caps for low object integrity, poor composition, and low semantic render quality.

The action space is bounded and discrete. It includes lighting and environment actions such as increasing key light strength, rotating the environment, increasing environment strength, and increasing contact shadow; object actions such as small vertical lift/lower operations, yaw correction, and scale changes; and material actions such as value, saturation, hue, and roughness adjustment. The controller chooses actions either from VLM-suggested actions or from deterministic fallback mappings from issue tags and diagnostic fields. The implementation also contains safety logic: actions are skipped near parameter bounds, repeated sign flips can freeze a target parameter, and no-op is returned when no safe action is available.

The following algorithm is the minimum paper-level description needed to make the inner loop operationally reproducible without exposing every implementation detail.

\begin{table}[t]
\centering
\small
\begin{tabular}{p{0.18\linewidth}p{0.72\linewidth}}
\toprule
Component & Operational definition \\
\midrule
Input & Current render, mask and render metadata, optional previous render, optional object reference, optional pseudo-reference insertion image, current control state, review schema, and bounded action space. \\
Review & VLM scores lighting, object integrity, composition, render-quality semantics, and overall quality; VLM also emits issue tags, suggested actions, diagnostics, pairwise preference, and asset viability. CV scores exposure, sharpness, mask/framing when available, and programmatic grounding/contact consistency. \\
Action selection & Prefer valid VLM-suggested actions. If unavailable, map diagnostics and issue tags to bounded actions such as lighting adjustment, exposure/environment correction, object scale and position correction, support-plane grounding, material adjustment, render filtering, regeneration, or script/config edits. \\
Safety & Skip frozen or bound-saturated parameters, prevent oscillatory sign flips, keep task-locked variables fixed, and route missing meshes or failed renders to rejection or replacement. \\
Stopping & Stop when the review verdict is keep/pass, when the asset should be abandoned, when a plateau is detected, when no safe action remains, or when the maximum round cap is reached. \\
Log & Persist per-round control state, render paths, aggregate review JSON, VLM trace text, selected actions, score fields, issue tags, and the final routing decision. \\
\bottomrule
\end{tabular}
\caption{Operational contract for the generation-time inner loop.}
\label{tab:inner-loop-contract}
\end{table}

\paragraph{Algorithm sketch.}
For each object, initialize the control state from the scene template and task constraints. Render the current object in the fixed scene. Review the render with the VLM/CV reviewer and compute the hybrid score and route. If the route is pass or the freeform verdict is keep, persist the current state as the accepted canonical state. If the asset is judged non-viable, mark it for regeneration or rejection. Otherwise, select a bounded action from the VLM suggestion list or from the diagnostic fallback map, apply the action to the control state, and rerender. The loop terminates when the sample is accepted, rejected, plateaued, or reaches the round cap. This description is sufficient to expose the operational contract used by the internal quality gate in the ablation chain.

\FloatBarrier
\subsection{Supplementary External-Comparison Audit Figures}

These audit views correspond to the reader-facing external comparison in the main text: \textbf{Base}, \textbf{Public LoRA}, and \textbf{Ours +DualGate}. Their role is diagnostic rather than headline-setting. They show that the workflow-built scene-aware rotation dataset does not help only on aggregate means; it also produces a stronger eight-azimuth rotation probe under normalized summaries and angle-level breakdowns. Throughout these audit figures, the eight evaluation bins are labeled as 45\(^\circ\), 90\(^\circ\), \ldots, 360\(^\circ\), where 360\(^\circ\) denotes the front-equivalent full-rotation evaluation slot.

\subsubsection{SpatialEdit-Bench Audit Figures}

\begin{figure}[!tbp]
\centering
\begin{subfigure}[t]{0.32\linewidth}
\centering
\includegraphics[width=\linewidth]{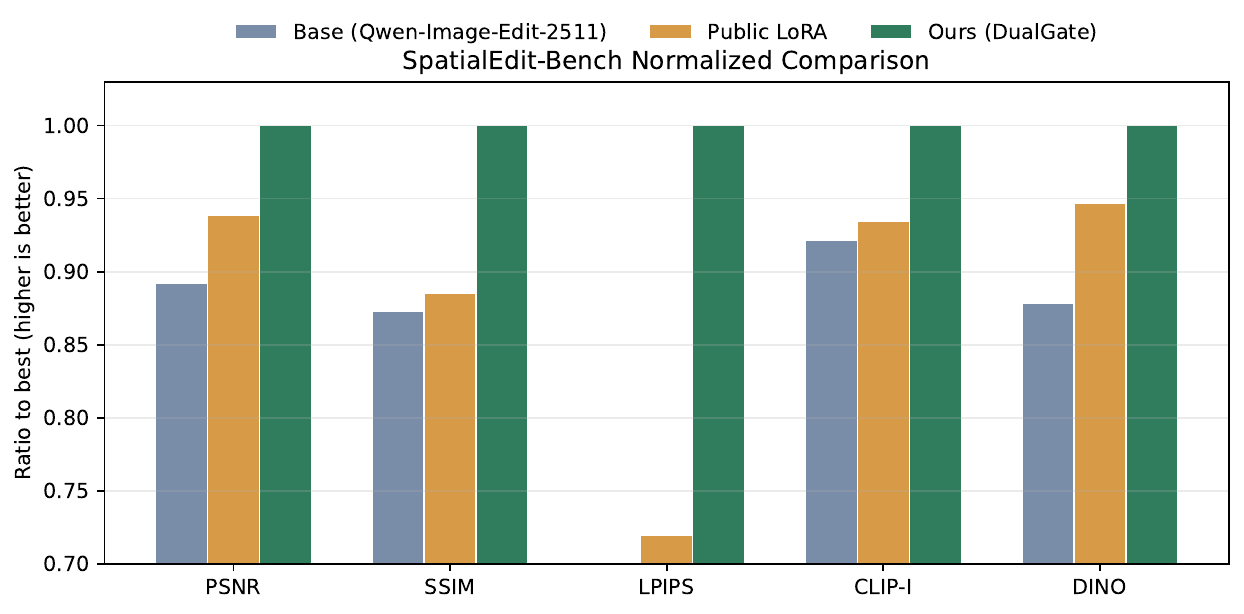}
\caption{Normalized comparison.}
\label{fig:external-spatialedit-normalized-audit}
\end{subfigure}
\hfill
\begin{subfigure}[t]{0.32\linewidth}
\centering
\includegraphics[width=\linewidth]{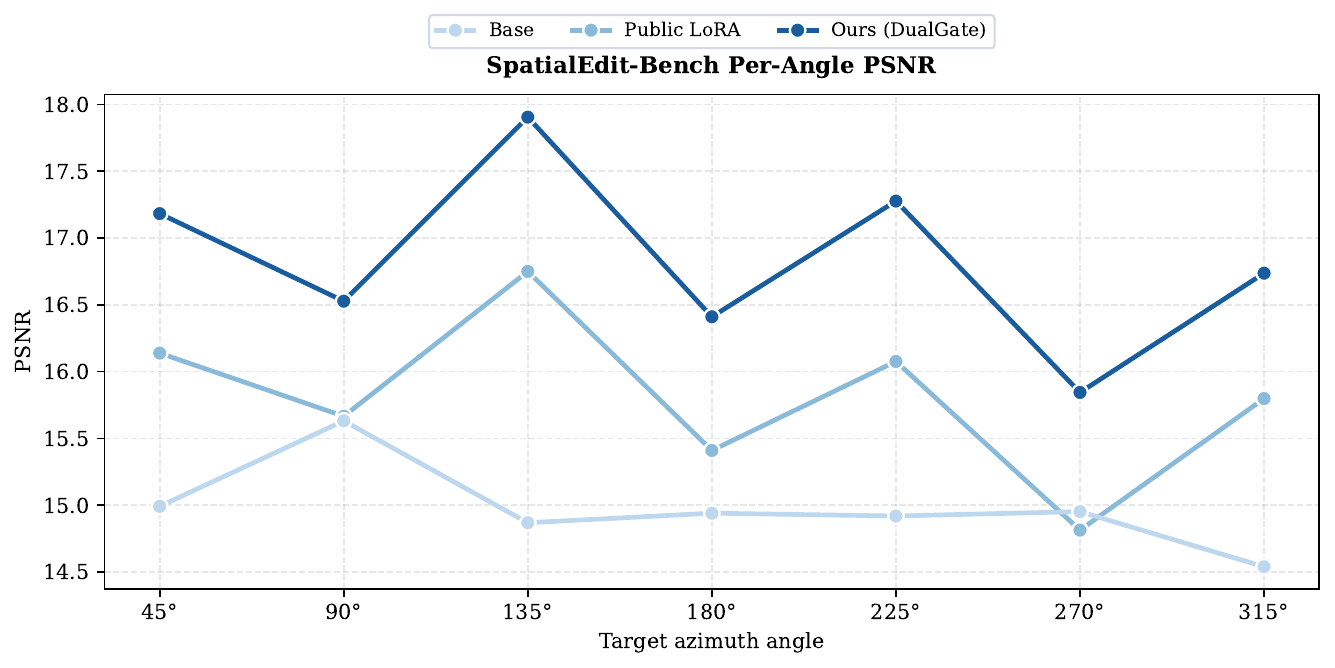}
\caption{Per-angle PSNR.}
\label{fig:external-spatialedit-per-angle-psnr-audit}
\end{subfigure}
\hfill
\begin{subfigure}[t]{0.32\linewidth}
\centering
\includegraphics[width=\linewidth]{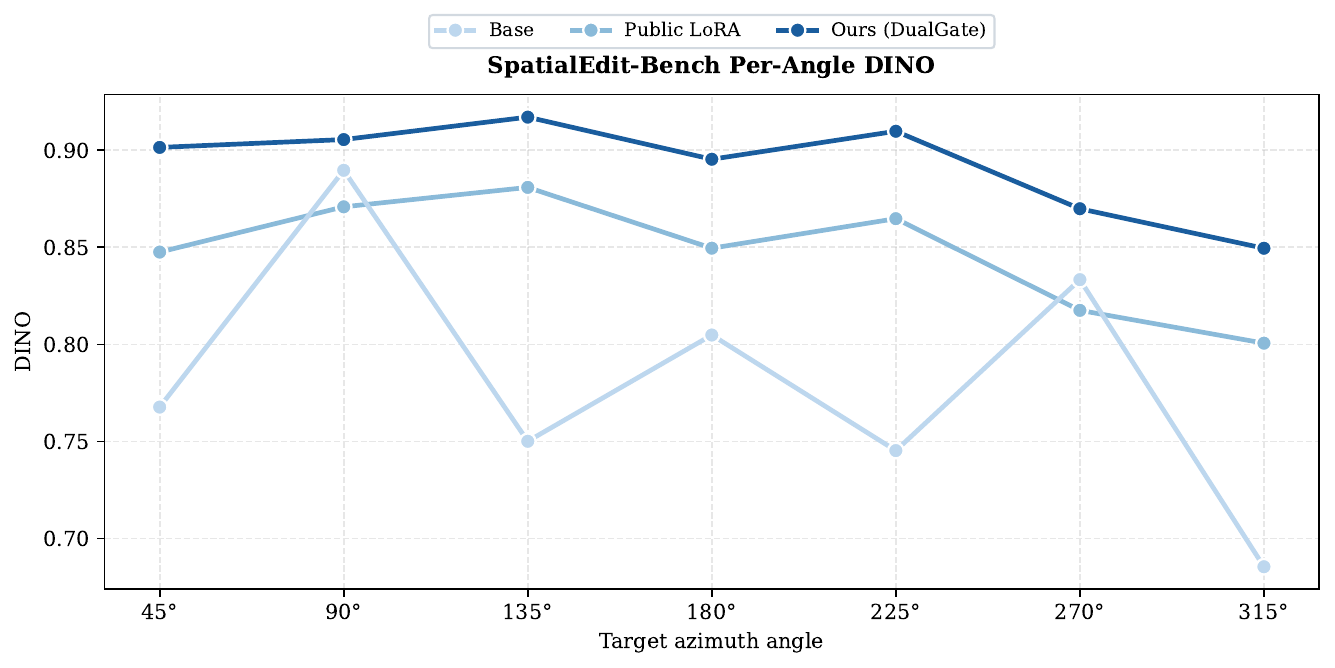}
\caption{Per-angle DINO.}
\label{fig:external-spatialedit-per-angle-dino-audit}
\end{subfigure}
\caption{Supplementary audit views for the \spatialedit{} external comparison. Positive values in the normalized subplot indicate improvement after accounting for whether a metric is higher-is-better or lower-is-better; the per-angle PSNR and DINO subplots show that the same ordering remains visible across angle bins.}
\end{figure}

\subsubsection{Eval1 Test Set Audit Figures}

\begin{figure}[!tbp]
\centering
\begin{subfigure}[t]{0.32\linewidth}
\centering
\includegraphics[width=\linewidth]{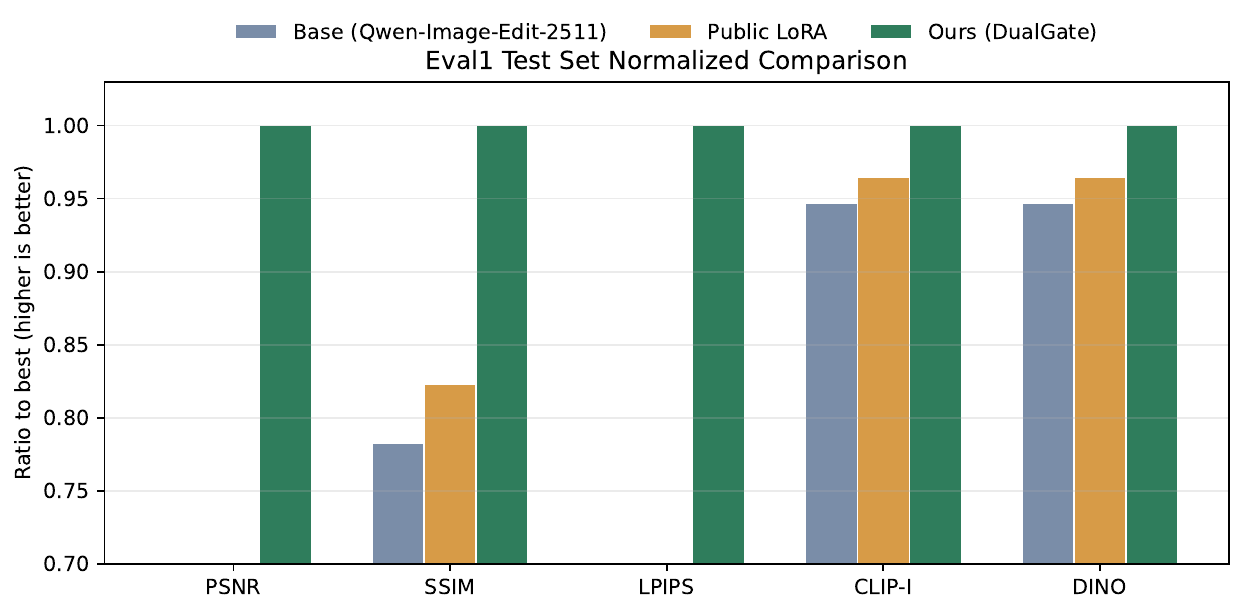}
\caption{Normalized comparison.}
\label{fig:external-testset-normalized-audit}
\end{subfigure}
\hfill
\begin{subfigure}[t]{0.32\linewidth}
\centering
\includegraphics[width=\linewidth]{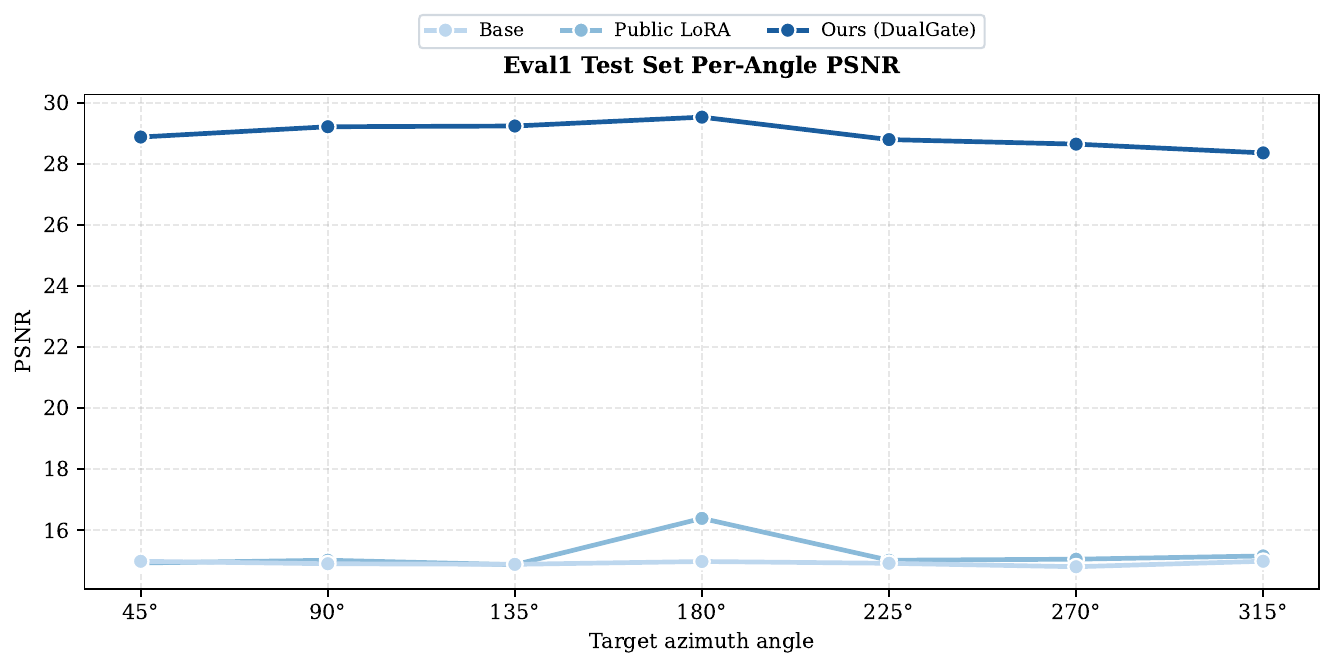}
\caption{Per-angle PSNR.}
\label{fig:external-testset-per-angle-psnr-audit}
\end{subfigure}
\hfill
\begin{subfigure}[t]{0.32\linewidth}
\centering
\includegraphics[width=\linewidth]{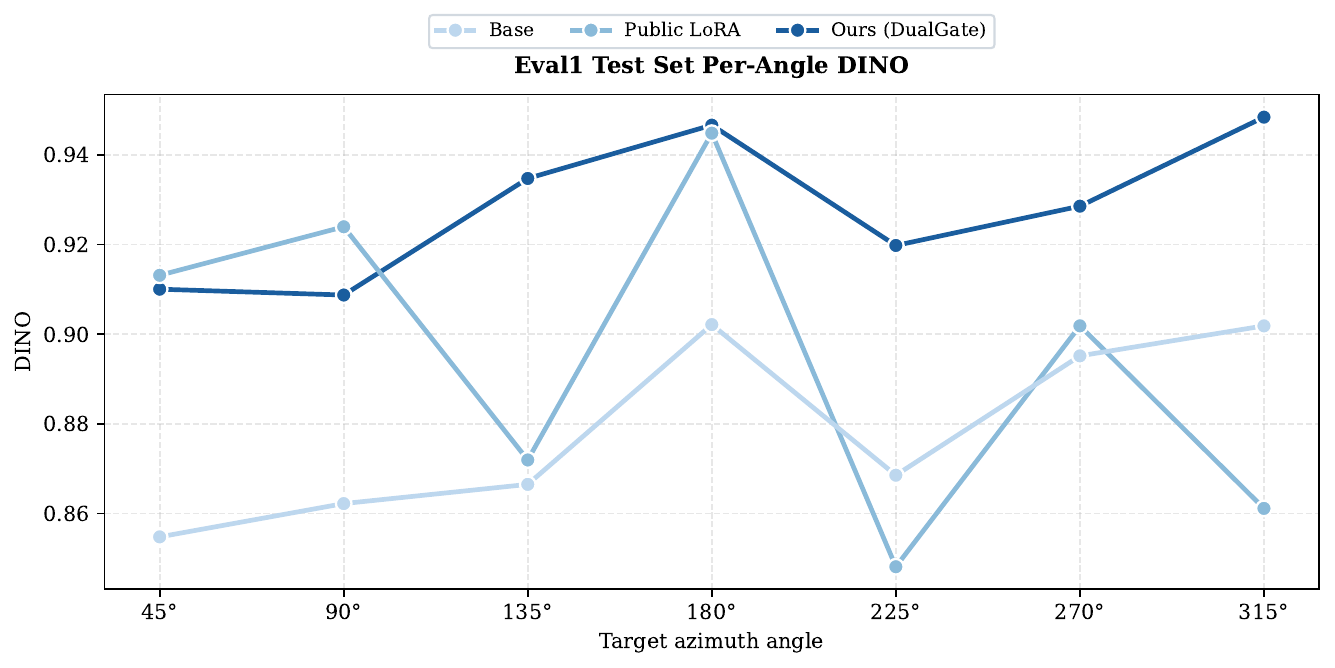}
\caption{Per-angle DINO.}
\label{fig:external-testset-per-angle-dino-audit}
\end{subfigure}
\caption{Supplementary audit views for the Eval1 Test Set external comparison. Positive values in the normalized subplot indicate improvement after accounting for metric direction; the per-angle subplots show that the same model ordering persists across the angle breakdown.}
\end{figure}

\FloatBarrier
\subsection{Supplementary Ablation Audit Figures}

The appendix adds three supplementary views of the same four-stage chain reported in the main text. These figures are not alternative headline results; they are audit views that make the consistency of the trend easier to inspect across metric families.

\begin{figure}[!tbp]
\centering
\begin{subfigure}[t]{0.32\linewidth}
\centering
\includegraphics[width=\linewidth]{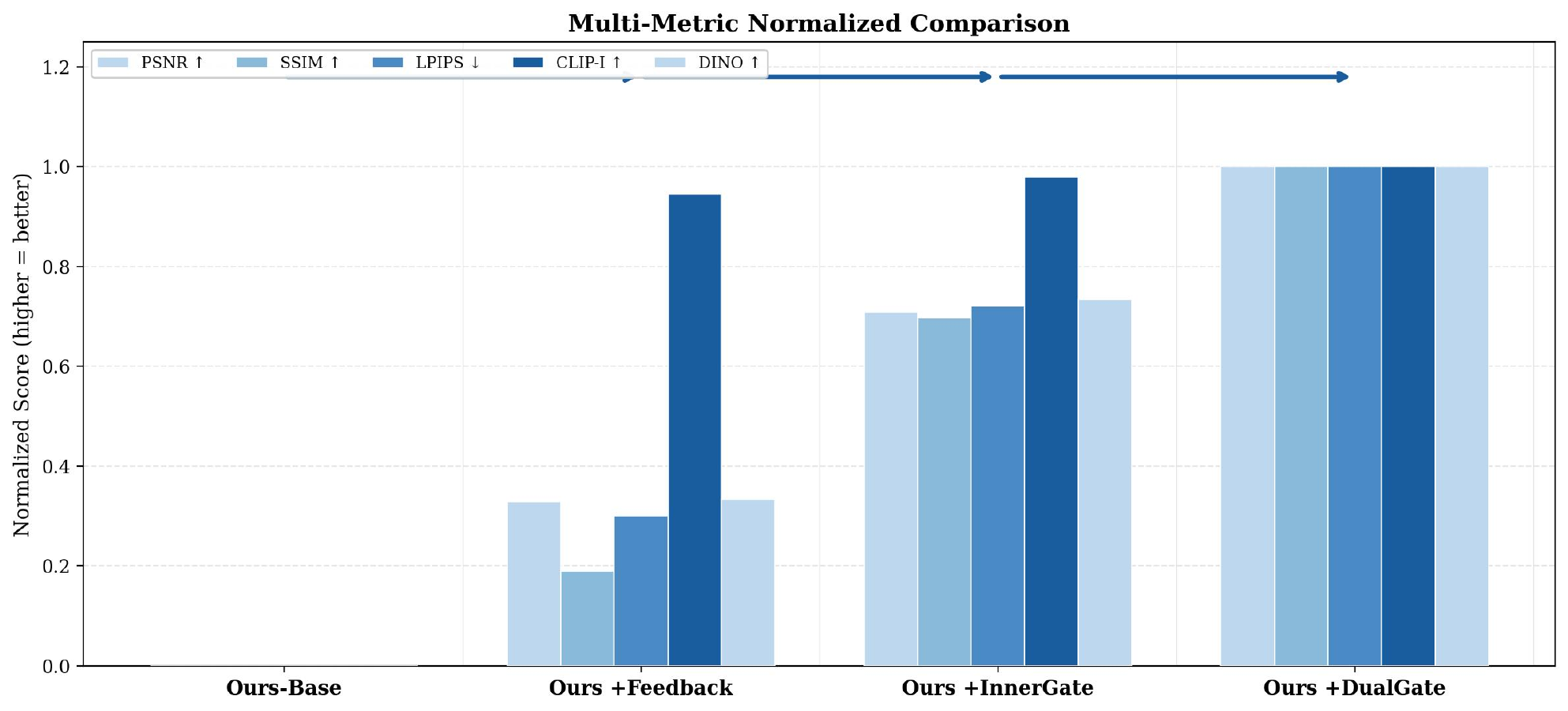}
\caption{Normalized comparison.}
\label{fig:ablation-normalized-audit}
\end{subfigure}
\hfill
\begin{subfigure}[t]{0.32\linewidth}
\centering
\includegraphics[width=\linewidth]{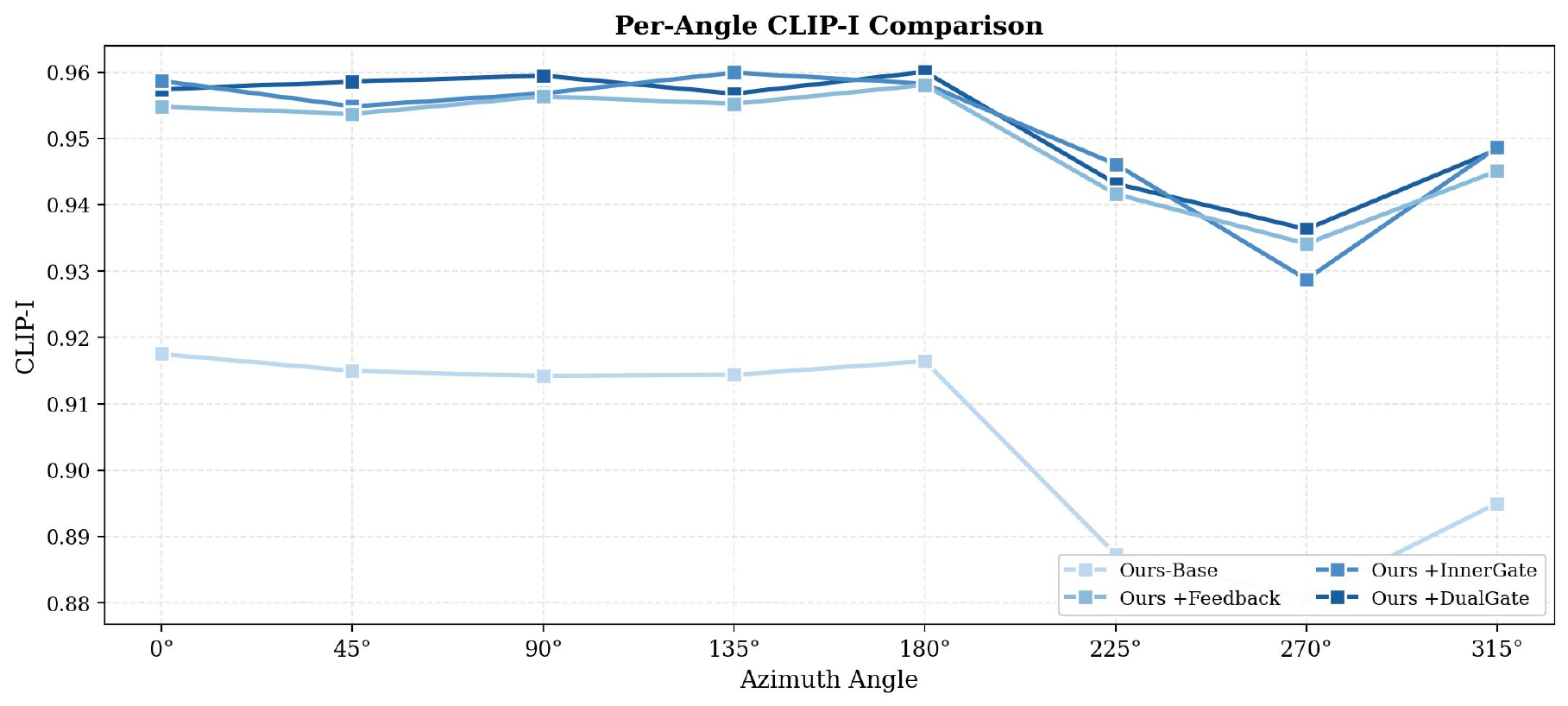}
\caption{Per-angle CLIP-I.}
\label{fig:ablation-clipi-audit}
\end{subfigure}
\hfill
\begin{subfigure}[t]{0.32\linewidth}
\centering
\includegraphics[width=\linewidth]{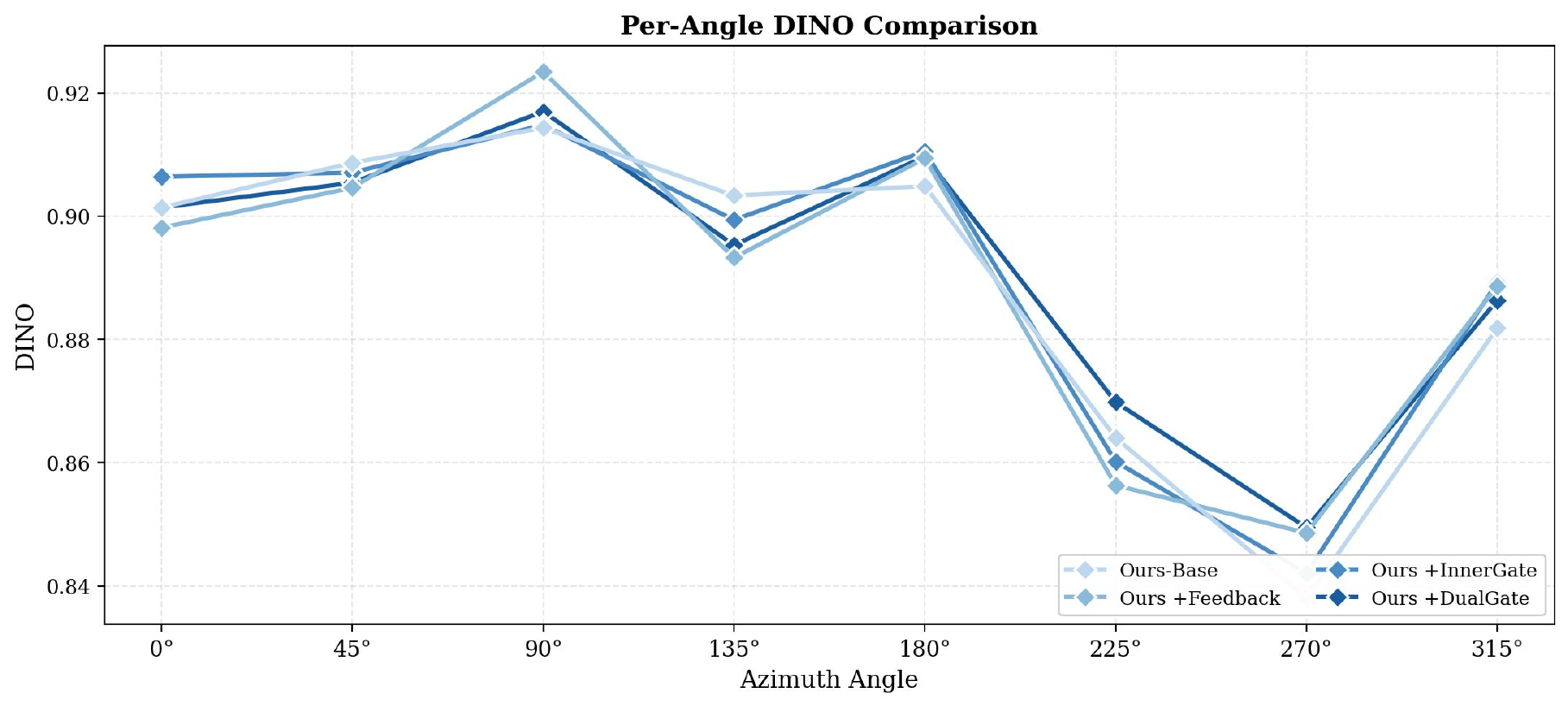}
\caption{Per-angle DINO.}
\label{fig:ablation-dino-audit}
\end{subfigure}
\caption{Supplementary audit views for the four-stage ablation chain. The normalized subplot summarizes direction-adjusted improvements, while the CLIP-I and DINO subplots show that semantic and identity-sensitive gains remain visible across angle subgroups.}
\end{figure}

\bibliographystyle{plainnat}
\bibliography{bib/references}

\begin{thebibliography}{31}
\providecommand{\natexlab}[1]{#1}
\providecommand{\url}[1]{\texttt{#1}}
\expandafter\ifx\csname urlstyle\endcsname\relax
  \providecommand{\doi}[1]{doi: #1}\else
  \providecommand{\doi}{doi: \begingroup \urlstyle{rm}\Url}\fi

\bibitem[Almohammadi et~al.(2025)Almohammadi, Mikaeili, Nag, Hassanpour, Tagliasacchi, and Mahdavi-Amiri]{almohammadi2025cora}
Amirhossein Almohammadi, Aryan Mikaeili, Sauradip Nag, Negar Hassanpour, Andrea Tagliasacchi, and Ali Mahdavi-Amiri.
\newblock {Cora}: Correspondence-aware image editing using few step diffusion, 2025.
\newblock URL \url{https://arxiv.org/abs/2505.23907}.

\bibitem[Bahmani et~al.(2025)]{bahmani2025lyra}
Sherwin Bahmani et~al.
\newblock {Lyra}: Generative {3D} scene reconstruction via video diffusion model self-distillation, 2025.
\newblock URL \url{https://arxiv.org/abs/2509.19296}.

\bibitem[Bar-On et~al.(2025)Bar-On, Cohen-Bar, and Cohen-Or]{baron2025editp23}
Roi Bar-On, Dana Cohen-Bar, and Daniel Cohen-Or.
\newblock {EditP23}: {3D} editing via propagation of image prompts to multi-view, 2025.
\newblock URL \url{https://arxiv.org/abs/2506.20652}.

\bibitem[{Blender Foundation}(2026)]{blenderFoundationBlender}
{Blender Foundation}.
\newblock Blender: Free and open source {3D} creation software.
\newblock \url{https://www.blender.org/}, 2026.
\newblock Accessed as an implementation tool reference.

\bibitem[Cao et~al.(2025)Cao, Chen, Pan, and Liu]{cao2025physx}
Ziang Cao, Zhaoxi Chen, Liang Pan, and Ziwei Liu.
\newblock {PhysX}: Physical-grounded {3D} asset generation, 2025.
\newblock URL \url{https://arxiv.org/abs/2507.12465}.

\bibitem[Carion et~al.(2025)Carion, Gustafson, Hu, Debnath, Hu, Suris, Ryali, Alwala, Khedr, Huang, Lei, Ma, Guo, Kalla, Marks, Greer, Wang, Sun, R{"a}dle, Afouras, Mavroudi, Xu, Wu, Zhou, Momeni, Hazra, Ding, Vaze, Porcher, Li, Li, Kamath, Cheng, Doll{'a}r, Ravi, Saenko, Zhang, and Feichtenhofer]{carion2025sam3segmentconcepts}
Nicolas Carion, Laura Gustafson, Yuan-Ting Hu, Shoubhik Debnath, Ronghang Hu, Didac Suris, Chaitanya Ryali, Kalyan~Vasudev Alwala, Haitham Khedr, Andrew Huang, Jie Lei, Tengyu Ma, Baishan Guo, Arpit Kalla, Markus Marks, Joseph Greer, Meng Wang, Peize Sun, Roman R{"a}dle, Triantafyllos Afouras, Effrosyni Mavroudi, Katherine Xu, Tsung-Han Wu, Yu~Zhou, Liliane Momeni, Rishi Hazra, Shuangrui Ding, Sagar Vaze, Francois Porcher, Feng Li, Siyuan Li, Aishwarya Kamath, Ho~Kei Cheng, Piotr Doll{'a}r, Nikhila Ravi, Kate Saenko, Pengchuan Zhang, and Christoph Feichtenhofer.
\newblock {SAM 3}: Segment anything with concepts, 2025.
\newblock URL \url{https://arxiv.org/abs/2511.16719}.

\bibitem[Cvejic et~al.(2025)Cvejic, Eldesokey, and Wonka]{cvejic2025partedit}
Aleksandar Cvejic, Abdelrahman Eldesokey, and Peter Wonka.
\newblock {PartEdit}: Fine-grained image editing using pre-trained diffusion models, 2025.
\newblock URL \url{https://arxiv.org/abs/2502.04050}.

\bibitem[{fal}(2026)]{fal2026qwenmultianglelora}
{fal}.
\newblock {Qwen-Image-Edit-2511-Multiple-Angles-LoRA}.
\newblock \url{https://huggingface.co/fal/Qwen-Image-Edit-2511-Multiple-Angles-LoRA}, 2026.
\newblock Hugging Face model card.

\bibitem[Fang et~al.(2025)Fang, Li, Liang, Zheng, Mao, Liu, Tang, Zhou, and Tan]{fang2025spatialgen}
Chuan Fang, Heng Li, Yixun Liang, Jia Zheng, Yongsen Mao, Yuan Liu, Rui Tang, Zihan Zhou, and Ping Tan.
\newblock {SPATIALGEN}: Layout-guided {3D} indoor scene generation, 2025.
\newblock URL \url{https://arxiv.org/abs/2509.14981}.

\bibitem[Feng et~al.(2025)]{feng2025seed3d}
Jiashi Feng et~al.
\newblock {Seed3D 1.0}: From images to high-fidelity simulation-ready {3D} assets, 2025.
\newblock URL \url{https://arxiv.org/abs/2510.19944}.

\bibitem[Han et~al.(2025)]{han2025unireditbench}
Feng Han et~al.
\newblock {UniREditBench}: A unified reasoning-based image editing benchmark, 2025.
\newblock URL \url{https://arxiv.org/abs/2511.01295}.

\bibitem[Hu et~al.(2025)Hu, Liu, Tan, Yang, and Wang]{hu2025imageeditingprograms}
Yujia Hu, Songhua Liu, Zhenxiong Tan, Xingyi Yang, and Xinchao Wang.
\newblock Image editing as programs with diffusion models, 2025.
\newblock URL \url{https://arxiv.org/abs/2506.04158}.

\bibitem[Lai et~al.(2025)]{lai2025hunyuan3d25}
Zeqiang Lai et~al.
\newblock {Hunyuan3D 2.5}: Towards high-fidelity {3D} assets generation with ultimate details, 2025.
\newblock URL \url{https://arxiv.org/abs/2506.16504}.

\bibitem[Lei et~al.(2025)]{lei2025hunyuan3dstudio}
Biwen Lei et~al.
\newblock {Hunyuan3D Studio}: End-to-end {AI} pipeline for game-ready {3D} asset generation, 2025.
\newblock URL \url{https://arxiv.org/abs/2509.12815}.

\bibitem[Li et~al.(2025{\natexlab{a}})Li, Gu, Liu, Lin, Wei, Grover, and Kuen]{li2025lavidao}
Shufan Li, Jiuxiang Gu, Kangning Liu, Zhe Lin, Zijun Wei, Aditya Grover, and Jason Kuen.
\newblock {Lavida-O}: Elastic large masked diffusion models for unified multimodal understanding and generation, 2025{\natexlab{a}}.
\newblock URL \url{https://arxiv.org/abs/2509.19244}.

\bibitem[Li et~al.(2025{\natexlab{b}})]{li2025step1x3d}
Weiyu Li et~al.
\newblock {Step1X-3D}: Towards high-fidelity and controllable generation of textured {3D} assets, 2025{\natexlab{b}}.
\newblock URL \url{https://arxiv.org/abs/2505.07747}.

\bibitem[Liu et~al.(2025)]{liu2025step1xedit}
Shiyu Liu et~al.
\newblock {Step1X-Edit}: A practical framework for general image editing, 2025.
\newblock URL \url{https://arxiv.org/abs/2504.17761}.

\bibitem[Pu et~al.(2025)]{pu2025picabench}
Yuandong Pu et~al.
\newblock {PICABench}: How far are we from physically realistic image editing?, 2025.
\newblock URL \url{https://arxiv.org/abs/2510.17681}.

\bibitem[Qian et~al.(2025)]{qian2025picobanana400k}
Yusu Qian et~al.
\newblock {Pico-Banana-400K}: A large-scale dataset for text-guided image editing, 2025.
\newblock URL \url{https://arxiv.org/abs/2510.19808}.

\bibitem[{Qwen Team}(2025)]{qwen2025qwenimageedit2511}
{Qwen Team}.
\newblock {Qwen-Image-Edit-2511}.
\newblock \url{https://huggingface.co/Qwen/Qwen-Image-Edit-2511}, 2025.
\newblock Hugging Face model card.

\bibitem[Rampini et~al.(2025)Rampini, Madan, Roy, Zamani, and Cheung]{rampini2025texturemapping}
Arianna Rampini, Kanika Madan, Bruno Roy, AmirHossein Zamani, and Derek Cheung.
\newblock A scalable attention-based approach for image-to-{3D} texture mapping, 2025.
\newblock URL \url{https://arxiv.org/abs/2509.05131}.

\bibitem[Schouten et~al.(2025)Schouten, Kaya, Belongie, and Papadopoulos]{schouten2025poem}
Marco Schouten, Mehmet~Onurcan Kaya, Serge Belongie, and Dim~P. Papadopoulos.
\newblock {POEM}: Precise object-level editing via {MLLM} control, 2025.
\newblock URL \url{https://arxiv.org/abs/2504.08111}.

\bibitem[Tang et~al.(2025)Tang, Li, and Fan]{tang2025zeroscene}
Xiang Tang, Ruotong Li, and Xiaopeng Fan.
\newblock {ZeroScene}: A zero-shot framework for {3D} scene generation from a single image and controllable texture editing, 2025.
\newblock URL \url{https://arxiv.org/abs/2509.23607}.

\bibitem[{Team Hunyuan3D}(2025)]{teamhunyuan3d2025hunyuan3d21}
{Team Hunyuan3D}.
\newblock {Hunyuan3D 2.1}: From images to high-fidelity {3D} assets with production-ready {PBR} material, 2025.
\newblock URL \url{https://arxiv.org/abs/2506.15442}.

\bibitem[Wu et~al.(2025)Wu, Li, Zhou, Lin, Gao, Yan, Yin, Bai, Xu, Chen, Chen, Tang, Zhang, Wang, Yang, Yu, Cheng, Liu, Li, Zhang, Meng, Wei, Ni, Chen, Cao, Peng, Qu, Wu, Wang, Yu, Wen, Feng, Xu, Wang, Zhang, Zhu, Wu, Cai, and Liu]{wu2025qwenimage}
Chenfei Wu, Jiahao Li, Jingren Zhou, Junyang Lin, Kaiyuan Gao, Kun Yan, Sheng-ming Yin, Shuai Bai, Xiao Xu, Yilei Chen, Yuxiang Chen, Zecheng Tang, Zekai Zhang, Zhengyi Wang, An~Yang, Bowen Yu, Chen Cheng, Dayiheng Liu, Deqing Li, Hang Zhang, Hao Meng, Hu~Wei, Jingyuan Ni, Kai Chen, Kuan Cao, Liang Peng, Lin Qu, Minggang Wu, Peng Wang, Shuting Yu, Tingkun Wen, Wensen Feng, Xiaoxiao Xu, Yi~Wang, Yichang Zhang, Yongqiang Zhu, Yujia Wu, Yuxuan Cai, and Zenan Liu.
\newblock {Qwen-Image} technical report, 2025.
\newblock URL \url{https://arxiv.org/abs/2508.02324}.

\bibitem[Xiao et~al.(2026)Xiao, Zhang, Song, Chen, Li, Jiang, Ren, Lin, Huang, Huang, Li, Duan, and Qi]{xiao2026spatialedit}
Yicheng Xiao, Wenhu Zhang, Lin Song, Yukang Chen, Wenbo Li, Nan Jiang, Tianhe Ren, Haokun Lin, Wei Huang, Haoyang Huang, Xiu Li, Nan Duan, and Xiaojuan Qi.
\newblock {SpatialEdit}: Benchmarking fine-grained image spatial editing, 2026.
\newblock URL \url{https://arxiv.org/abs/2604.04911}.

\bibitem[Yang et~al.(2025)Yang, Lin, Xu, Li, and Chen]{yang2025highfidelity25dlatents}
Xin Yang, Jiantao Lin, Yingjie Xu, Haodong Li, and Yingcong Chen.
\newblock Advancing high-fidelity {3D} and texture generation with {2.5D} latents, 2025.
\newblock URL \url{https://arxiv.org/abs/2505.21050}.

\bibitem[Ye et~al.(2025)]{ye2025imgedit}
Yang Ye et~al.
\newblock {ImgEdit}: A unified image editing dataset and benchmark, 2025.
\newblock URL \url{https://arxiv.org/abs/2505.20275}.

\bibitem[Yu et~al.(2026)]{yu2026i2e}
Jinghan Yu et~al.
\newblock {I2E}: From image pixels to actionable interactive environments for text-guided image editing, 2026.
\newblock URL \url{https://arxiv.org/abs/2601.03741}.

\bibitem[Zhu et~al.(2025)Zhu, Li, Feng, Wu, Qiao, and Yuan]{zhu2025georemover}
Zixin Zhu, Haoxiang Li, Xuelu Feng, He~Wu, Chunming Qiao, and Junsong Yuan.
\newblock {GeoRemover}: Removing objects and their causal visual artifacts, 2025.
\newblock URL \url{https://arxiv.org/abs/2509.18538}.

\bibitem[Zou et~al.(2025)]{zou2025beyondtextualcot}
Zhentao Zou et~al.
\newblock Beyond textual {CoT}: Interleaved text-image chains with deep confidence reasoning for image editing, 2025.
\newblock URL \url{https://arxiv.org/abs/2510.08157}.

\end{thebibliography}

\end{document}